\documentclass[preprint,12pt,nopreprintline]{elsarticle}

\usepackage{amssymb}
\usepackage{amsmath}

\usepackage{amsmath,amssymb,amsfonts}
\usepackage{algorithmic}
\usepackage{algorithm}
\usepackage{soul, color, xcolor}
\usepackage{setspace}

\usepackage{graphicx}
\usepackage{epstopdf}
\usepackage{textcomp}
\usepackage{subcaption}
\usepackage{url}
\usepackage{bm}
\usepackage{caption,setspace}
\usepackage{newtxmath}
\usepackage{multirow}
\usepackage{geometry}
\usepackage{enumitem}

\setlist[enumerate]{listparindent=\parindent, parsep=0pt}
\begin{document}

\begin{frontmatter}



\title{Hierarchical Deep Deterministic Policy Gradient for Autonomous Maze Navigation of Mobile Robots}


\author[labela]{Wenjie Hu} 
\ead{huenjie@student.usm.my}

\author[labela]{Ye Zhou\corref{cor1}}
\ead{zhouye@usm.my}

\author[labela]{Hann Woei Ho}
\ead{aehannwoei@usm.my}

\affiliation[labela]
            {
            addressline={School of Aerospace Engineering, Engineering Campus, Universiti Sains Malaysia, 14300 Nibong Tebal, Pulau Pinang, Malaysia}}

\cortext[cor1]{Corresponding author}

\begin{abstract}
Maze navigation is a fundamental challenge in robotics, requiring agents to traverse complex environments efficiently. While the Deep Deterministic Policy Gradient (DDPG) algorithm excels in control tasks, its performance in maze navigation suffers from sparse rewards, inefficient exploration, and long-horizon planning difficulties, often leading to low success rates and average rewards, sometimes even failing to achieve effective navigation. To address these limitations, this paper proposes an efficient Hierarchical DDPG (HDDPG) algorithm, which includes high-level and low-level policies. The high-level policy employs an advanced DDPG framework to generate intermediate subgoals from a long-term perspective and on a higher temporal scale. The low-level policy, also powered by the improved DDPG algorithm, generates primitive actions by observing current states and following the subgoal assigned by the high-level policy. The proposed method enhances stability with off-policy correction, refining subgoal assignments by relabeling historical experiences. Additionally, adaptive parameter space noise is utilized to improve exploration, and a reshaped intrinsic-extrinsic reward function is employed to boost learning efficiency. Further optimizations, including gradient clipping and Xavier initialization, are employed to improve robustness. The proposed algorithm is rigorously evaluated through numerical simulation experiments executed using the Robot Operating System (ROS) and Gazebo. Regarding the three distinct final targets in autonomous maze navigation tasks, HDDPG significantly overcomes the limitations of standard DDPG and its variants, improving the success rate by at least 56.59\% and boosting the average reward by a minimum of 519.03 compared to baseline algorithms.
\end{abstract}




\end{frontmatter}



\section{Introduction}
\label{sec:introduction}
In recent years, autonomous navigation for mobile robots has emerged as a pivotal research focus, driven by its potential to enhance efficiency and safety across diverse applications,  which mainly strive to enable mobile robots to reach the target point while effectively avoiding obstacles \cite{zhu2017target,liu2023graph}. 
Maze navigation, as one of the diverse autonomous navigation tasks, poses distinct challenges. It demands an efficient long-horizon path finding strategy in the absence of a comprehensive global map, requiring the agent to learn the optimal sequence of actions to navigate in the maze with a higher success rate.
This problem is particularly relevant in critical scenarios such as warehouse logistics, search and rescue operations in disaster-stricken areas, and autonomous exploration of hazardous or unknown environments, where precise and adaptive path planning is crucial for success. As a classic case, most researchers have solved maze navigation by knowing the full maze map in advance\cite{kumar2017maze,barnouti2016pathfinding}, which is needed to build the global map of the maze in advance. However, this approach presents several drawbacks. Firstly, it requires prior knowledge of the entire environment, which is impractical in unknown or dynamic settings. Secondly, it limits the exploration ability of the agent, decreasing the adaptability to a new or changing environment. Lastly, it could be computationally intensive, especially for large or complex maze scenarios, posing challenges for memory-constrained systems \cite{bai2024path,zhu2021deep}. 

To solve these problems, mapless navigation approaches have proven to be much more capable for target-driven autonomous navigation, which involves determining a collision-free path to a specified goal, only relying on local environmental information without pre-existing environmental descriptions and global maps constructed\cite{taivirtual,dobrevski2021deep,zhelo2018curiosity}. 
A widespread approach to mapless navigation is learning-based navigation strategies. Reinforcement learning (RL) is a biologically inspired machine learning approach, which has been preferred for various robotic control applications due to its simplicity and flexibility to select actions that maximize accumulated reward during interactions with the environment\cite{zhou2022online}. However, a notable limitation of RL is that, during the early stages of training, the policy tends to be ineffective due to the agent's limited understanding of the environment. Additionally, substantial exploration and long training time caused by the `curse of dimensionality' may be required before the task is successfully executed, depending on its complexity \cite{kaelbling1996reinforcement,tizhoosh2006opposition}.
Deep learning (DL) is a branch of machine learning that mimics how the human brain works and allows computational models to learn and understand representations of data directly by multiple processing layers without manually designing intricate processing steps \cite{lecun2015deep,sharifani2023machine}. DL models are highly capable of handling high-dimensional data and efficiently mitigating the challenges posed by the `curse of dimensionality' \cite{han2018solving,hutzenthaler2020proof}. DL has achieved transformative progress in a wide range of fields, including autonomous systems and robotics, computer vision, as well as speech processing and recognition.

In recent years, deep reinforcement learning (DRL), integrating DL and RL, has succeeded in addressing a variety of complex decision-making challenges that were once beyond the capabilities of machines, such as autonomous vehicles \cite{kiran2021deep}, robots controlling and manipulations \cite{lillicrap2015continuous,carlucho2020adaptive}, and video games \cite{lample2017playing}. For autonomous navigation of the mobile robot, DRL empowers the mobile robot to autonomously learn from interactions with its environment, making optimal policies without human-crafted features or explicit maps. By utilizing deep learning's ability to automatically abstract features from raw input data, DRL is particularly advantageous for mapless autonomous navigation. DRL enables mobile robots to dynamically adapt to their surroundings and make real-time decisions based on raw sensor data. The ability to learn in an end-to-end manner is highly beneficial in situations where pre-existing maps are not available, or when mobile robots navigate in unfamiliar or dynamic environments. By deriving policies directly from environmental interactions, DRL-based mapless autonomous navigation algorithms can effectively deal with complex nonlinear relationships between sensor data and actions while mitigating the challenges posed by high-dimensional spaces \cite{li2023deep}. To directly deal with continuous action spaces, the Deep Deterministic Policy Gradient (DDPG) algorithm was introduced, a model-free algorithm based on the deterministic policy gradient and actor-critic structure. DDPG leverages both a policy network as the actor and a value network as the critic to learn optimal policies in continuous action spaces \cite{grondman2012survey,costa2023ac2cd}. This actor-critic framework is well-suited for learning in continuous action spaces, making it particularly capable and advantageous for robotic control tasks. In the context of autonomous maze navigation for mobile robots, the actor policy determines the linear and angular velocities in a continuous action space based on the given state input and was optimized by maximizing the expected cumulative reward. The critic policy estimates the value of the actions selected in a specific state and is trained by minimizing the corresponding loss function.

However, the traditional DDPG algorithm encounters several significant challenges when applied to autonomous maze navigation tasks. One of the primary issues is the inefficiency in learning, which arises due to the trial-and-error nature of DRL. This inefficiency becomes particularly pronounced in environments where rewards are sparse or delayed, making it difficult for the agent to effectively learn from limited feedback. Consequently, DDPG typically requires a substantial number of interactions with the environment to converge toward optimal policies, leading to high computational costs and long training times \cite{9409758}. Moreover, slow convergence presents another critical hurdle in autonomous maze navigation, where the agent must explore and learn long-term path planning, but the traditional DDPG algorithm struggles to efficiently converge to the optimal solution, especially in complex maze environments \cite{zhelo2018curiosity}. This results in suboptimal performance, with low success rates and average scores. In addition, the traditional DDPG algorithm requires effective exploration and exploitation policies to balance obstacle avoidance and target achievement. However, it is a big challenge to adjust the weights of the factors, especially in complicate autonomous maze navigation tasks \cite{zhu2022hierarchical}. Initially, we tried to optimize the DDPG algorithm with double critics and double actors, called D$_4$PG, to generate a more accurate target value estimation, and improved sampling efficiency \cite{10753589,tseng2022autonomous}. However, the improvement did not lead to significant breakthroughs in autonomous maze navigation tasks.

Subsequently, we leveraged a hierarchical advanced DDPG algorithm, which incorporates individual high-level and low-level policies \cite{wohlke2021hierarchies, mannucci2016hierarchical}. Specifically, the high-level policy is implemented by an improved DDPG algorithm. It generates intermediate subgoals from a longer-horizon perspective and on a higher temporal scale based on the spatial relationship between the current position of the mobile robot and the ultimate goal. High-level policy provides a favorable, collision-free direction and long-term path planning, helping the mobile robot avoid collisions while approaching the final destination. Then, the low-level policy also leverages an advanced DDPG algorithm to generate the movements of the agent interacting with the environment by observing a certain state, and the subgoal assigned by the high-level policy. After taking an action at the low level, the high-level controller provides an intrinsic reward to the low-level controller \cite{nachum2018data}. The hierarchical structure improves learning efficiency by simplifying complex, long-horizon maze navigation tasks. The high-level policy plans strategic paths, while the low-level policy handles precise motion controls, enabling efficient decision-making. By assigning meaningful subgoals, the high-level policy directs exploration, reducing redundancy, and improving success rates. 

However, a significant challenge arises from the off-policy experience buffer in the hierarchical DDPG algorithm. As training progresses, the low-level policy may evolves and improves. 
Consequently, even when provided with the same high-level subgoal, the current low-level policy may generate actions that differ from those produced by the previous low-level policy. The historical experiences in the off-policy experience buffer no longer accurately reflect what the current low-level policy would achieve, which may lead to inaccurate value estimates, ineffective sub-goal assignments, unstable training, etc. To resolve this inconsistency, we introduced an off-policy correction method that relabels the subgoals in the high-level off-policy experience buffer so that they can align with the current low-level policy. Specifically, the subgoal in the past high-level experience is relabeled to be consistent with the observed action sequence, making it more likely to have occurred under the current low-level policy. Additionally, adaptive parameter space noise is incorporated into both high-level and low-level policies to enhance exploration. This approach improves the balance between exploration and exploitation, while also increasing training efficiency and stability by introducing adaptive random perturbations to the actor network parameters. Additionally, the noise magnitude is dynamically adjusted based on the agent's learning progress \cite{plappert2017parameter}.
Furthermore, an innovative and effective target-driven intrinsic and extrinsic reward function is utilized to enhance learning efficiency. In addition, to improve the stability and resilience, practical enhancements, such as the gradient clipping technique and Xavier initialization, are applied.

The main contributions of this paper are described as follows: 
\begin{enumerate}
\item {A hierarchical DDPG algorithm is leveraged for autonomous maze navigation of the mobile robot, where the high-level policy generates intermediate subgoals for strategic planning, and the low-level policy produces specific control actions according to the environment observation and the assigned subgoal. This structure effectively simplifies complicate, long-horizon maze navigation tasks, enhancing learning efficiency.}
\item {An off-policy correction mechanism is utilized to resolve the inconsistency issue in the off-policy experience buffer of the hierarchical DDPG algorithm, leading to more accurate value estimation, effective subgoal assignment, and stable training.}
\item {Adaptive parameter space noise is applied to both high-level and low-level policies to facilitate exploration, leading to a better balance between exploration and exploitation, as well as improved learning efficiency and stability.}
\item {Innovative and effective target-oriented reshaping intrinsic and extrinsic reward functions are employed, which promotes effective exploration by guiding the agent toward final goal avoiding collision, thus reducing unnecessary exploration, and improving success rates.}
\end{enumerate}

The remaining sections of the paper are organized as follows. Section 2 presents the problem statement and relevant research. Section 3 details the methodology of the proposed algorithm. Section 4 describes the specific implementation of the proposed algorithm in autonomous maze navigation tasks for the mobile robot. Section 5 outlines the experimental setup and compares the results of the proposed algorithm with those of the DDPG and its variant. Finally, Section 6 concludes this paper by summarizing key findings and suggesting directions for future research.
\section{Fundamentals}

\subsection{Problem Statement}
Autonomous maze navigation is a fundamental challenge in robotics and autonomous systems, requiring a mobile agent to traverse complex environments and reach a predefined goal without collisions \cite{devo2020deep}. The DDPG algorithm, a prominent DRL method, has demonstrated considerable efficiency in robotic control problems. However, its use in autonomous maze navigation remains constrained by a low success rate and average score, primarily due to difficulty in obtaining meaningful rewards, suboptimal exploration strategies, challenges in managing long-horizon path planning, and limited visibility. To address these challenges, we sought to utilize D$_4$PG, an advanced DDPG algorithm, incorporating dual critic and actor networks. This approach aimed to ensure more precise target value estimation, and mitigate suboptimal exploration. However, it resulted in minimal performance improvement. 

To tackle these challenges, we proposed a hierarchical DRL algorithm called HDDPG, which consists of high-level and low-level controllers, implemented using an advanced variant of the DDPG algorithm. The proposed method enables more efficient long-horizon path planning, improves decision-making efficiency, and enhances learning stability while maintaining scalability for complex autonomous maze navigation tasks \cite{zhou2019hybrid}. Furthermore, adaptive parameter space noise is applied to both high-level and low-level actor networks, enabling an efficient exploration policy that facilitates a better balance between exploration and exploitation. This approach accelerates the learning process and mitigates the risk of converging to local optima. Additionally, novel target-driven reshaped intrinsic and extrinsic reward functions are incorporated to speed up learning and improve convergence efficiency.

\subsection{Deep Reinforcement Learning Foundation}

 Recently, RL is a well-established discipline, typically positioned within the realm of machine learning. Additionally, due to its emphasis on learning behaviors, it shares significant links with various other domains, including psychology, operations research, mathematical optimization, and more \cite{wiering2012reinforcement}. The agent learns an optimal policy through trial and error, guided by environment observations and quantitative reward signals derived from its previous actions, which is modeled as the Markov Decision Process (MDP). In RL, the agent encounters a sequential decision-making problem, where at each time step $t$, it perceives the current state $s_t$ from a state space $S$. Based on this perception, the agent takes an action $a_t$ generated by the policy $\pi(s, \theta^{\pi})$ from the action space $A$. Upon executing action $a_t$, the agent receives a scalar reward $r_t$ from the environment, and the state transitions from $s_t$ to the next state $s_{t+1}$ with a certain probability $p_t\in P$. In RL, the tuple ($s_t$, $a_t$, $r_t$, $s_{t+1}$) is known as an experience.

Deep reinforcement learning (DRL) integrates reinforcement learning (RL) with deep learning (DL) techniques by leveraging deep neural networks to approximate key functions in RL. These include the state value networks, denoted as $V(s,\theta^V)$, the state-action value networks $Q(s,a;\theta^Q)$, and the policy network $\pi(s,\theta^{\pi})$, where $\theta^V$, $\theta^Q$, and $\theta^{\pi}$ 
represent the neural network parameters, which are updated via the stochastic gradient method. Consequently, DRL algorithms are generally classified into two main categories: value-based and policy-based approaches. Deep $Q$ network (DQN), the pioneering value-based DRL algorithm, is inherently restricted to discrete action spaces because it computes the value of each action, necessitating the discretization of continuous actions. This process can lead to reduced precision and increased computational demands. Conversely, policy-based DRL methods directly optimize the policy without requiring discretization, making them more effective for continuous action spaces or environments with high-dimensional state representations.  Policies can be deterministic by assigning a specific action to each state, or stochastic, where actions are selected based on a probability distribution. One prominent example of a policy-based DRL algorithm, tailored for continuous action spaces, is the DDPG algorithm. DDPG is widely recognized for its simplicity of implementation, adaptability to high-dimensional state spaces, and stable convergence in various robot control applications.

\subsection{Deep Deterministic Policy Gradient and Its Variant}
 DDPG algorithm employs the actor-critic (AC) architecture, comprising a policy-based function approximation neural network $\pi$, known as the actor, and a value-based function approximation neural network $Q$, known as the critic. The actor neural network $\pi$ learns a deterministic policy by mapping a given state to an action, and the critic neural network $Q$ evaluates the policy using a state-action value network, leading to efficient policy optimization. Specifically, the actor network $\pi(s,\theta^{\pi})$ generates a deterministic action $a_t$ for a given state $s_t$ at time step $t$. Meanwhile, the critic network $Q(s,a;\theta^Q)$ evaluates the selected action $a_t$ under the specific state $s_t$. Once the agent executes the action $a_t$, it receives a reward $r_t$ from the environment, and the state transitions to the subsequent state $s_{t+1}$. This interaction results in an experience tuple ($s_t$, $a_t$, $r_t$, $s_{t+1}$), which is sequentially stored in the experience replay buffer for future training.

DDPG is an off-policy DRL approach, in which the actor and critic networks are trained off-policy using past experiences stored in the experience replay buffer. This strategy is able to help alleviate sample correlations and improve training efficiency. During training, a mini-batch of experience sequences is randomly drawn from the experience replay buffer to update both the critic and actor networks. 
The critic network is optimized through gradient descent to minimize the loss between the predicted $Q$ value and the target $Q$ value. Furthermore, to improve stability and robustness, DDPG incorporates a target critic network $Q'(s,a;\theta^{Q'})$ and a target actor network $\pi'(s,\theta^{\pi'})$, guaranteeing consistent target $Q$ value estimation throughout training, and reducing dependence on updates from the online networks to mitigate estimation bias.

The loss function $L(\theta^Q)$ for the critic network is determined using the online network $Q(s,a;\theta^{Q})$ and the target network $Q'(s,a;\theta^{Q'})$. It can be expressed as follows:
 \begin{align}                      
 L(\theta^Q) &\approx\frac{1}{K}\sum_{t=1}^{K} (y_t-Q(s_t,a_t;\theta^{Q}))^2,
 \label{e:Q}
 \end{align}
where $K$ represents the number of sampled experiences, and $\theta^{Q}$ denotes the parameters of the online critic network. The action $a_t$ is selected as the optimal action by the online actor network for the given environment state $s_t$ at time step $t$, and $y_t$ denotes the target $Q$ value at time step $t$, expressed as follows:
\begin{align}
y_t=\begin{cases}
  r_t& \text{if $s_{t+1}$ is the end} \\
  r_t+\gamma Q'(s_{t+1},\pi'(s_{t+1},\theta^{\pi'});\theta^{Q'})& \text{otherwise},
\end{cases}
     \end{align}
where $\theta^{Q'}$ and $\theta^{\pi'}$ refer to the parameters of the target critic network and target actor network, and $r_t$ is the reward received from the environment after performing the specific action $a_t$. The discount factor $\gamma$ determines the balance between immediate and future rewards, and $\gamma$ is set to 0.99 in this study. Therefore, the parameter $\theta^Q$ of the online critic is optimized using the gradient descent approach as follows:
\begin{align}
\theta^Q_{t+1}={\theta^Q_t}-\eta\nabla_{\theta^Q|_{\theta^Q=\theta^Q_t}}{L(\theta^Q)},
\label{e:gd}
\end{align}
where \textit{$\eta$} $\in$ [0,1], is the learning rate, controlling the step size for updating the network parameters.

The actor network $\pi(s,\theta^{\pi})$ is optimized by maximizing the expected cumulative future rewards. Additionally, its loss function $L(\theta^\pi)$ and the corresponding gradient $\nabla_{\theta^\pi}{L(\theta^\pi)}$ can be respectively expressed as follows:
\begin{align}
 L\left(\theta^{\pi}\right) &= \mathbb{E}_{s \sim \mathcal{D}}\left[Q\left(s, \pi\left(s,\theta^{\pi}\right); \theta^{Q}\right)\right]\label{e:e8}, \\
 \nabla_{\theta^\pi}{L(\theta^\pi)} &\approx  \frac{1}{K}\sum_{t=1}^{K} \nabla_{a_t}Q(s_t,a_t;\theta^Q)\nabla_{\theta^\pi}\pi(s_t,\theta^\pi), \label{e:e9}
 \end{align}
where $\mathbb{E}_{s\sim\mathcal{D}}$ represents the expected value of the $Q$ value, with states $s$ are sampled from the experience replay buffer $\mathbb{D}$. The term $K$ means the number of  experiences sampled from the experience replay buffer $\mathbb{D}$, and the parameter $\theta^\pi$ of the actor network is updated  using the gradient ascent strategy, which is expressed as follows:
\begin{align}
\theta^\pi_{t+1}=\theta^\pi_{t}+\eta\nabla_{\theta^\pi|_{\theta^\pi=\theta^\pi_t}}{J(\theta^\pi)}.
\label{e:ga}
\end{align}

\begin{figure}[h]
\begin{center}
\includegraphics[width=1\linewidth]{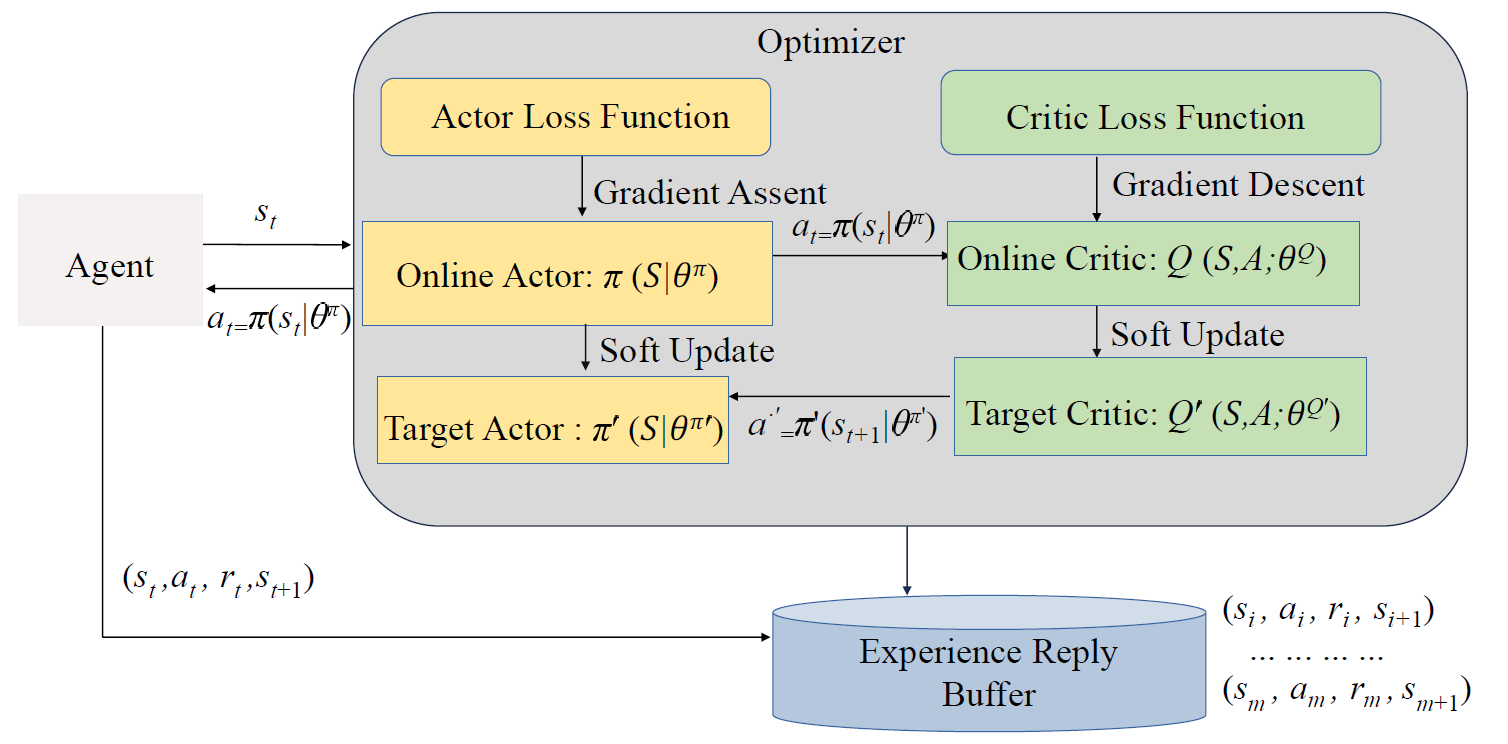}
\caption{At time step $t$, the online actor network $\pi(s,\theta^{\pi})$ generates an action 
$a_t$, based on the current state $s_t$. After performing the action $a_t$, the environment provides a reward $r_t$, transitioning to the next state $s_{t+1}$. The experience tuple 
($s_t$, $a_t$, $r_t$, $s_{t+1}$) is stored in the experience replay buffer. During training, mini-batches of experience are sampled to update the online critic and actor networks. 
}
\label{fig:DDPG}
\end{center}              
\end{figure}

Target networks $Q'(s,a;\theta^{Q'})$ and $\pi'(s,\theta^{\pi'})$ have the same architecture as their respective online networks. The parameters $\theta^{Q'}$ and $\theta^{\pi'}$ are optimized using a soft update mechanism. It is a commonly used technique in DRL, which applies gradual updates by maintaining a small, constant proportional relationship with the parameters of the online network, controlled by $\tau$. This approach helps reduce the frequency of parameter updates and improves training stability \cite{9448360}. The smooth modification of parameters $\theta^{Q'}$ and $\theta^{\pi'}$ is carried out as follows:
\begin{align} 
\theta^{Q'}=(1-\tau)\theta^{Q'}+\tau\theta^{Q},\\
\theta^{\pi'}=(1-\tau)\theta^{\pi'}+\tau\theta^{\pi},
\end{align}
where $\tau$ determines the smoothness of the soft update process, and setting $\tau$ to 1 reduces the soft update to a traditional hard update. In this study, $\tau$ is configured as 0.005. The general architecture of the DDPG algorithm is illustrated in Fig. \ref{fig:DDPG}. 

The DDPG algorithm has demonstrated remarkable performance in autonomous control tasks within continuous action spaces. However, its application to autonomous maze navigation for mobile robots faces significant challenges with low success rates and average scores, primarily due to sparse rewards and inefficient exploration. To address these challenges, we introduced double critics and double actors into the DDPG framework, resulting in a variant algorithm called D$_4$PG. This modification helps reduce overestimation and improves the accuracy of target $Q$ value estimation. Although the D$_4$PG algorithm achieves approximately a 32.56\% improvement in success rate for autonomous maze navigation scenarios when the final target is relatively close to the starting point, its performance shows only minimal improvement compared to the original DDPG algorithm when the target is further away. These limitations of both DDPG and D$_4$PG underscore the need for more robust approaches that can better handle the challenges posed by intricate maze navigation environments. One such approach is Hierarchical Deep Reinforcement Learning (HDRL), which has proven to be an effective framework for addressing the complicate decision-making applications, including autonomous navigation \cite{wohlke2021hierarchies,bischoff2013hierarchicaL}.
\section{Hierarchical DDPG with off-policy correction and adaptive parameter space noise Method}

\subsection{Hierarchical DRL Framework}
HDRL is regarded as a promising framework, due to its inherent ability to break down a long-term task into a sequence of simple and elementary sub-tasks, which can be learned using DRL. For autonomous maze navigation tasks, planning directly from the start to the ultimate goal in complex maze environments can be challenging due to the vast search space and numerous possible paths, which is more likely to result in sub-optimal strategies with low success rates and average scores. To address these challenges, we employ a hierarchical advanced DDPG algorithm to train a mobile robot for autonomous maze navigation, achieving significant improvements in both success rate and average score compared to the original DDPG algorithm and its variant D$_4$PG. 

In our approach, the proposed algorithm is organized into two distinct layers: a high-level policy and a low-level policy. The high-level (HL) policy determines subgoals for long-term navigation corresponding to the ultimate goal and the current position of the mobile robot. For the agent, the subgoals generated by the high-level policy are closer and more achievable compared to the final goal. Additionally, a series of subgoals provides the mobile robot with guidance on movement direction, including obstacle avoidance and optimal path selection. The low-level policy generates specific actions based on the current environmental state and the subgoal established by the high-level policy, guiding the mobile robot toward the designated subgoals \cite{staroverov2020real}. The mobile robot then interacts with the environment, performing specific actions that yield intrinsic rewards, which are used to refine its behavior. By simultaneously utilizing both high-level and low-level policies, our approach simplifies the task of autonomous maze navigation for the mobile robot. This hierarchical dual-policy strategy reduces task complexity, enhances exploration efficiency, and improves overall learning performance.

\subsection{High level DRL}
The proposed algorithm directly utilizes the state observations obtained from the environment by the sensor. The high-level action $a^h$, is defined as the subgoal, which is an intermediate target location that guides the mobile robot toward the final goal. The subgoal $a^h$ is defined as $a^h=[a^h_x, a^h_y]$, representing x-axis and y-axis coordinates. This approach ensures that our HDRL framework is entirely general and eliminates the need for manual configurations or further multi-task designs. The high-level updates the subgoal over longer time intervals. DDPG is well-suited for high-level decision-making due to its ability to handle continuous action spaces, which aligns with the need to generate precise subgoals in autonomous maze navigation tasks. Its deterministic policy guarantees stable and controllable subgoal generation, which is pivotal for effective path planning. Additionally, the DDPG algorithm focuses on optimizing long-term cumulative expected rewards, making it ideal for long-term, high-level decision-making in intricate tasks.  \cite{wang2018deep,sehgal2022aacher}. 

Specifically, at each time step $t$, the sensor obtains the current position $p_t$ of the mobile robot. The high-level policy $\pi^{h}$ produces an action $a^h_t\in\mathbb{R}^{d_p}$ based on $p_t$ and the final goal $f_g$. Here, $d_p$ represents the dimension of the mobile robot's position, set as 2 in this work. Subsequently, the low-level controller generates a deterministic action $a_t^{l}$ according to the subgoal $a^h_t$, and the position of the mobile robot may transit to $p_{t+1}$ after performing the action $a_t^{l}$. When the mobile robot is guided by the low-level controller from its current position to the subgoal $a^h_t$, a sequence of actions $a_t^{l}$,$a_{t+1}^{l}$......$a_{t+c}^{l}$ is generated. Here, $c$ represents the number of time steps required to reach the subgoal, but it is not a fixed scalar value.
Additionally, the tuple ($p_t$, $a^h_t$, $R_t$, $p_{t+c}$, $a_{t:t+c}$) is stored in the high-level experience replay buffer, which is used to optimize the high-level networks.

The high-level controller generates a new subgoal either when the mobile robot reaches the current subgoal or when the maximum time step limit for locating a subgoal is reached. A sequence of these subgoals constitutes the optimal path for the mobile robot to navigate. An essential challenge in this process is determining the appropriate distance around the robot's current position for subgoal generation. If this distance is set too small, the high-level module generates an excessive number of subgoals, increasing computational complexity and decision-making pressure. This can lead to the mobile robot following unnecessarily intricate or suboptimal paths. Additionally, frequent interactions between the high-level and low-level controllers may cause communication delays and consume unnecessary resources. Conversely, if subgoals are too sparse, the low-level controller is tasked with overly complex navigation challenges. This can hinder effective decision-making, reduce exploration efficiency, and result in navigation errors or a lower success rate. 

\subsection{Low Level DRL}

In the HDRL framework, the mobile robot does not navigate directly to the final goal but progresses through subgoals sequentially. The low-level controller is responsible for reaching these subgoals while avoiding collisions. It translates high-level directives into specific actions, such as adjusting movement and avoiding obstacles, acting as the bridge between strategic planning and physical actions. The DDPG algorithm is particularly well-suited for implementing the low-level policy in robotic navigation \cite{gao2023improved,jesus2019deep}.

\begin{figure*}[h]
    \centering
    \includegraphics[width=1\linewidth]{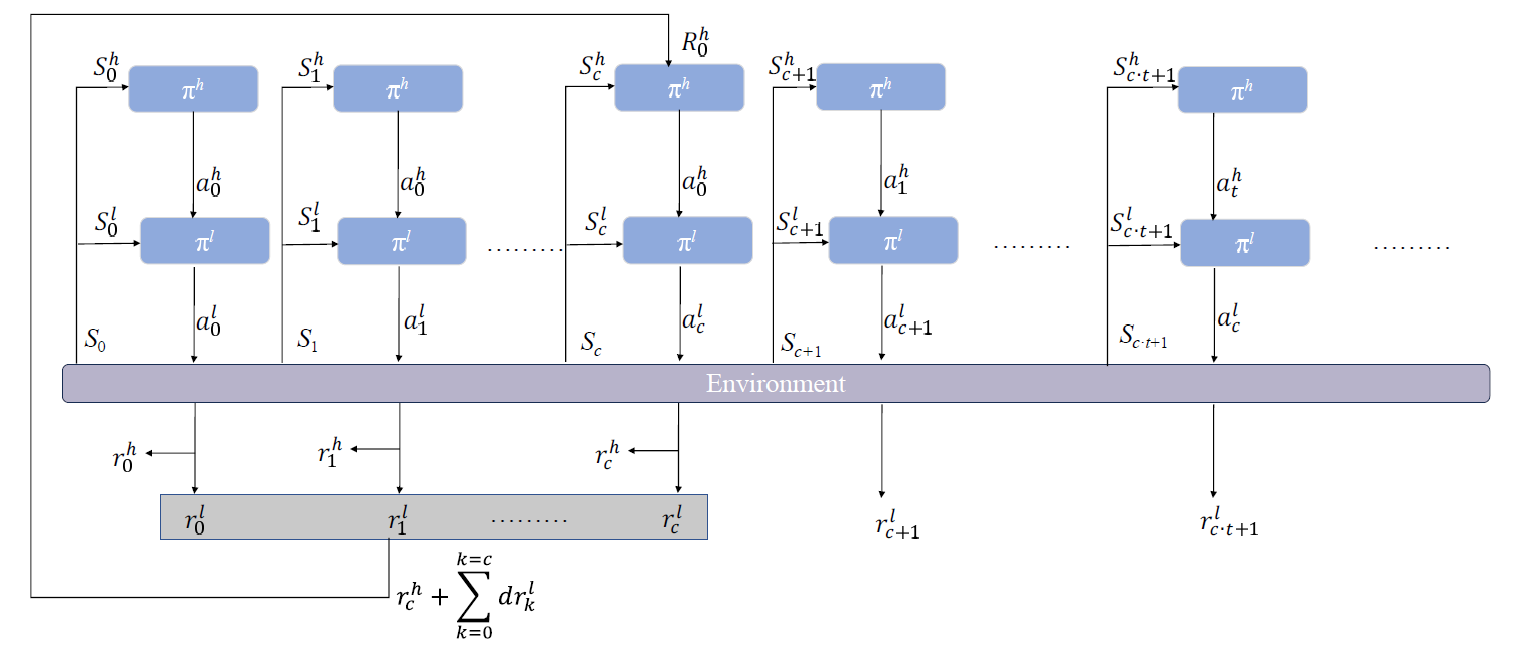}
    \caption{The hierarchical DRL framework, which consists of a high-level policy $\pi^h$  and a low-level policy $\pi^l$. At time step $t$, the high-level policy selects action $a^h_t$ based on the current position of the mobile robot and the final goal included in the high-level state $s_t^h$, guiding the agent’s strategic direction. The low-level policy takes its environmental observation $s_t^l$ and subgoal from the high-level policy to generate a primitive action $a_t^l$ to interact with the environment. After performing action $a_t^l$, low-level reward $r_t^l$ and high-level reward $r_t^h$ are obtained from the environment. Consequently, high-level and low-level states may transfer to the next new one. The high-level reward $r_t^h$ is calculated by combining cumulative low-level rewards and its achievement feedback. Once a subgoal is reached or a time step limit for finding a subgoal is exceeded, the high-level policy updates the subgoal.}
    \label{fig:hrl}
\end{figure*}

At time step $t$, the environment provides the current state information $s_t^l$ for the low-level policy. Based on $s_t^l$ and subgoal $a^h_t$, the low-level actor network $\pi^{l}$ generates a specific action $a_t^l$ including liner and angular velocities, represented as ${a_t^l}\sim\pi^{l}$($s_t^l$+$f_g$, $a^h_t$). Then the mobile robot executes the action $a_t^l$ in the maze scenario and receives an intrinsic reward $r_t^l$ from the environment. Subsequently, the low-level state transitions to the new state $s_{t+1}^l$, and an experience tuple ($s_t^l$, $a^h_t$, $a_t^l$, $r_t^l$, $s_{t+1}^l$) is stored in the experience reply buffer. The DDPG algorithm, being an off-policy method, allows the actor and critic networks to be trained using previously collected experiences. During training, a mini-batch of the experiences is randomly sampled from the buffer. The critic network in the low-level controller is leveraged to evaluate the state-action value $Q^l(a_t^l,s_t^l)$ after performing the action $a_t^l$ for a given state $s_t^l$. The target actor and critic networks are utilized to generate the target action ${a_t^l}'$ and compute the corresponding target state-action value $Q'^l({{a_t^l}',s_t^l})$. Subsequently, the critic network can be optimized by minimizing the loss between the expected $Q$ value $Q(a_t^l,s_t^l)$ and the target $Q$ value $Q'({{a_t^l}',s_t^l)}$ expressed in Eq. \eqref{e:Q}. 
Simultaneously, the actor network is updated by maximizing the expected $Q$ value over all sampled transitions, as shown in Equ. \eqref{e:e9}. Up to this, the hierarchical high-level and low-level framework in the work is depicted in Fig. \ref{fig:hrl}. In conclusion, the pseudocode for the proposed HDDPG algorithm is outlined in Algorithm \ref{alg:ap-hddpg}.

\subsection{Off-Policy Corrections for High-Level Training} 

Many previous studies have introduced two-level HDRL architectures incorporating subgoal design based on on-policy training. This dependency arises from the evolving behavior of the low-level policy may introduce non-stability for the high-level controller. Consequently, off-policy experiences may present different transitions even when designed on the same subgoals. However, for HDRL methods to be practical in real-world applications, they must be sample-efficient, which means that they can learn effective policies with fewer interactions with the environment. Off-policy algorithms, like Deep Q Network (DQN) and DDPG, are more sample efficient since they can reuse previous experiences stored in an experience replay buffer, making them better suited for applications where data collection is limited or time-consuming. However, the biggest challenge of using off-policy in the HDRL structure is the inconsistency between the previously collected experiences and the current low-level policy. Specifically, since the low-level policy is varying under the high-level policy, the current low-level policy may generate actions that differ from those produced by the previous low-level policy when given the same high-level subgoal. This discrepancy results in the situation that the historical experiences no longer accurately indicate the behavior of the current low-level policy. As a result, this mismatch can lead to imprecise value estimations, ineffective subgoal assignments, and instability during training. To mitigate this issue, we adopt an off-policy correction technique that relabels the subgoals in historical high-level experiences, aligning the observed action sequences more closely with the behavior of the current low-level policy $\pi^{l}$ \cite{nachum2018data}.

Specifically, the high-level action $a^h_t$, which originally generates the historical low-level behavior sequences $a_{t:t+c}^l\sim\pi^{l}(s_{t:t+c}^l,a^h_t)$, replaced with a new subgoal $\widetilde{a^h_t}$ which is more likely to produce the same low-level behavior sequences under the current low-level policy. This relabeling process modifies the high-level experience tuple ($p_t$,$a^h_t$,$R_t$,$p_{t+c}$,$a^l_{t:t+c}$) by substituting the original high-level action ${a^h_t}$ with an alternative action $\widetilde{a^h_t}$, selected to maximize the probability $\pi^{l}(a_{t:t+c}^l|s_{t:t+c}^l$, $\widetilde{a^h_t}$).  Since most RL algorithms employ stochastic exploration strategies, the behavior policy is inherently probabilistic. Consequently, the probability log $\pi^{l}(a_{t:t+c}^l|s_{t:t+c}^l$, $\widetilde{a^h_t}$) may be computed as follows:
\begin{align}
\log \pi^{\text{l}}\left(a_{t:t+c}^l \mid s_{t:t+c}^l, \tilde{a^h_t}\right)
&\propto -\frac{1}{2} \sum_{i=t}^{t+c} \big\| a_i^l - \pi^{\text{l}}\left(s_i^l, \tilde{a^h_i}\right) \big\|_2^2 + \text{const.},
\label{e:log}
\end{align}
where $\big\| x \big\|_2$ represents the Euclidean norm, which measures the deviation between the previous action stored in the low-level experience and the action generated by the current low-level policy.
To effectively approximate the maximization of this quantity, we compute this quantity for eight candidate goals randomly sampled from a Gaussian distribution centered at $p_{t+c} - p_t$. In addition, we include the original subgoal, $a^h_t$, and a subgoal related to the difference, $p_{t+c} - p_t$, in the candidate subgoal set, resulting in a total of ten candidates. This approach ensures a sufficiently diverse set of $\widetilde{a^h_t}$ to effectively approximate the probability described in Eq. \eqref{e:log}.
\subsection{Adaptive Parameter Space Noise} 

Although, the DDPG algorithm is proper to implement both low-level and high-level policies in the DRL framework for autonomous maze navigation, effective exploration strategies significantly enhance the performance of DRL in various applications. Its primary objective is to prevent agents from prematurely settling into suboptimal actions. However, how to efficiently explore the environment and collect informative experiences that could benefit policy learning toward the optimal one remains a formidable challenge in DRL \cite{10021988}. Facilitating an efficient exploration in autonomous maze navigation for mobile robots is complicate and difficult since mobile robots typically operate in high-dimensional and continuous action spaces with limited visibility, resulting in a vast exploration space and increasing the complexity of policy learning. 
In the conventional DDPG algorithm, random noise, either uncorrelated Gaussian noise or temporally correlated Ornstein-Uhlenbeck (OU) noise, is directly applied to the action output of the deterministic policy to enhance exploration. 

However, directly adding random noise to the actions leads to a stochastic and non-reproducible strategy, which can be unpredictable and inconsistent. It often results in inefficient exploration strategies, hindering the learning and training instability. Furthermore, even for a fixed state $s_t$, it is highly unlikely to produce the same action consistently because the noise added to the action space is entirely independent of the current state. To mitigate these challenges, an adaptive parameter space noise \cite{plappert2017parameter} is leveraged to high-level and low-level actor networks in our study. In this approach, the parameters of high-level and low-level actor networks are perturbed at the beginning of each episode and updated through gradient ascent. This ensures that the same action is taken whenever the same state is encountered, maintaining consistency. Additionally, the magnitude of this adaptive parameter space noise is adjusted dynamically based on how similar the perturbed policy's actions are to the original policy's actions. This ensures the noise remains effective throughout training.
Furthermore, the perturbation with the parameters of the actor network results in more complicated exploratory behavior, thereby accelerating the learning process and avoiding of premature convergence to a sub-optimal policy. 

Specifically, the adaptive scale of parameter space noise $\sigma$ is updated as follows :
\begin{align}
\sigma_{k+1}=\begin{cases}
  \alpha\sigma_k& \text{if $d$($\pi$, $ \overset{\sim}{\pi}$)$\le\varsigma$}\\
   \frac{1}{\alpha}\sigma_k & \text{otherwise},  
\end{cases}
     \end{align}
where $\alpha \in \mathbb{R}_{>0}$ is a scaling factor, setting $\alpha=1.01$ in our simulated experiments, and $\varsigma  \in \mathbb{R}_{>0}$ becomes a threshold value set as 0.2. In addition, $d$($\pi$, $\overset{\sim}{\pi}$) represents the distance between perturbed and unperturbed policies in action space, which can be generated as follows:
\begin{align}
d(\pi, \overset{\sim}{\pi})= \sqrt{\frac{1}{K}\sum_{i=1}^{K}\mathbb{E}_S[\pi(s)_i-\overset{\sim}{\pi}(s)_i]^2},
\label{e:14}
\end{align}
where $\mathbb{E}_S$[·] is estimated from a batch of states from the experience replay buffer and $K$ represents the number of sampled experience sequences.

\begin{algorithm}[H]
\footnotesize 
\caption{\textbf{Hierarchical Deep Deterministic
Policy Gradient with Off-policy Correction and
adaptive parameter space noise algorithm}}
\label{alg:ap-hddpg}
\begin{algorithmic}[1]
\STATE set the initial standard deviation $\sigma$ to 0.2, discount factor $\gamma$ to 0.99, learning rate $\eta$ to 0.001, etc.

\STATE Initialize high-level and low‑level critic networks $Q^h, Q^l$, actor networks $\pi^h,\pi^l$ and their target networks ${Q'}^h$, ${Q'}^l$, ${\pi'}^{h}$, ${\pi'}^{l}$
\STATE Initialize the perturbed high-level and low-level actor networks $\pi^h_p$, $\pi^l_p$ 
\STATE Initialize the parallel high-level and low-level experience replay buffers $A$ and $B$

\STATE Set $INITIAL$=True for initializing the high-level subgoal at the beginning.
\FOR{$episode = 1 \to N$}
\STATE Initialize environment state $s_0$, and final goal $f_g$
\STATE Hard update the perturbed high-level and low-level actor networks  $\pi^h_p$, $\pi^l_p$ with the online networks ${\pi^h}$, ${\pi^l}$
\STATE {Sample random noise $\varepsilon_{\omega}$, $\varepsilon_b$$\sim\mathcal{N}$(0,1)  }
\STATE \parbox[t]{\dimexpr\linewidth-\algorithmicindent}{Set the perturbed high-level and low-level actor networks $\pi^h_p$ and $\pi^l_p$ by adding noise $\sigma *\varepsilon_{\omega}$ to the weights, and $\sigma *\varepsilon_b$ to the biases of the online actor networks.}
\FOR{$t = 1 \to T$}
\IF {$INITIAL$}
   \STATE Calculate high-level action $a^h_t \gets$ $\pi^h_p$($s^h_t$, $f_g$; $\theta^{\pi^h_p}$)
   \STATE $INITIAL$=False
\ENDIF
\STATE Calculate low-level action $a^l_t \gets$ $\pi^l_p$($s^l_t$, $a^h_t$; $\theta^{\pi^l_p}$)

\STATE Perform action $a^l_t$\text{,} observe $r^l_t$, $s^l_{t+1}$, $s^h_{t+1}$
\STATE  \parbox[t]{\dimexpr\linewidth-\algorithmicindent} {Store low-level experience ($s^l_t$, $a^h_t$, $r^l_t$, $a^l_t$, $s^l_{t+1}$) in low-level experience replay buffer $B$}
\IF {reach to the current subgoal $a^h_t$ or do not reach the current subgoal within 100 steps }
   \STATE Calculate the next subgoal $a^h_{t+1} \gets$ $\pi^h_p$($s^h_{t+c}$,  $f_g$; $\theta^{\pi^h_p}$)
\ENDIF

\IF{Length of low-level experience replay buffer $\ge$ BATCH\_SIZE}
\STATE Sample mini-batch of $K$ transitions from low-level experience buffer $B$
\STATE Update $\theta_{Q^l}$ with gradient descent by minimizing the loss function in Eq. \ref{e:gd}.
\STATE Update $\theta_{\pi^l}$ with gradient assent by maximizing the loss function in Eq. \ref{e:ga}.
\STATE soft update target network ${Q'}^l$ and ${\pi'}^l$
\ENDIF
\IF{Length of high-level experience replay buffer $A$ $\ge$ BATCH\_SIZE}
\STATE Sample mini-batch of $K$ transitions $(s^h_t$, $f_g$, $a^h_t$, $r^h_t$, $s^h_{t+c}, a^l_{t:t+c})$ from high-level experience buffer $B$

\STATE replace the high-level action ${a^h_t}$ in the sampled transition with the alternative action $\widetilde{a^h_t}$, selected to maximize the probability $\pi^{l}(a^l_{t:t+c}|s^l_{t:t+c})$

\STATE Update $\theta_{Q^h}$ with gradient descent by minimizing the loss function in Eq. \ref{e:gd}.
\STATE Update $\theta_{\pi^h}$ with gradient assent by maximizing the loss function in Eq. \ref{e:ga}.
\STATE soft update target network ${Q'}^h$ and ${\pi'}^h$
\ENDIF

\ENDFOR
\STATE \parbox[t]{\dimexpr\linewidth-\algorithmicindent} {Sample all experience sequences from $A, B$ to calculate the difference $d_h$, and $d_l$}
\STATE \parbox[t]{\dimexpr\linewidth-\algorithmicindent} {Update the current standard derivations of the parameter space noise adding to high-level and low-level policies according to the difference  $d_h$, and $d_l$}
\ENDFOR

\end{algorithmic}
\end{algorithm}

\section{Autonomous Maze Navigation based on Hierarchical DDPG}

\subsection{Low-Level State Space Design}
The state represents the environmental information perceived by the agent, serving as a critical input for its decision-making process during training. For mobile robots to effectively navigate and avoid collisions in intricate environments, a well-designed and appropriate state space is essential. This state space must provide sufficient information for the mobile robot to accurately understand its surroundings and make optimal decisions. Simultaneously, the dimensionality and abstraction of the state space should be minimized to streamline the algorithm and reduce training time.

Our simulated experiments utilize the Waffle model of the TurtleBot3 mobile robot. The Waffle model features a laser sensor capable of detecting obstacles within a 360° field of view, measuring distances based on return time. The lidar sensor gathers data on the distances and relative angles between the mobile robot and nearby objects, while the odometer provides precise positional information about the robot. \cite{jesus2019deep,zhao2021dueling}. In this paper, the low-level state is defined as a 16-dimensional vector, structured as follows:
\begin{align}
    S=[{d_s}, l_p, a_p, \phi_0, d_g, f_g]\label{e:25},
\end{align}
where ${d_s}$ represents 10 equidistant measurements obtained by the lidar sensor within the range of -90° to 90°,  denoted as ${d_s}=[d_0,d_1,d_2......d_{9}]$. $l_p$ and $ a_p$ refer to the linear and angular velocity commands from the previous time step, respectively. $\phi_0$ indicates the angle of direction between the robot and the goal. $d_g$ represents the current distance to the subgoal. $f_g$ represents the positional information of the final goal, which is a 2-dimensional vector comprising the x-coordinate and y-coordinate.

\subsection{Low-Level Target-Driven Intrinsic Reward Function}

In DRL, the reward function plays a pivotal role, as it guides agent actions by assessing it effectiveness \cite{9341540}. Sparse rewards, which grant positive feedback only upon reaching the target or penalizing collisions, are simple and intuitive. However, in complex environments with dense or dynamic obstacles, the sparse rewards can be limited, often leading to the mobile robot being stuck or disoriented. Reward reshaping mitigates this issue by introducing intermediate rewards that offer more continuous and detailed feedback during task execution, thereby accelerating the learning process and enhancing the agent's task comprehension \cite{miranda2023generalization}. In the context of autonomous maze navigation for mobile robots, incorporating intrinsic reshaping rewards into the low-level controller is effective in enhancing its performance and adaptability.


 In the low-level controller, we propose a target-driven reshaping intrinsic reward function to formulate non-sparse rewards for the autonomous maze navigation of mobile robots. 
 The reward function consists of four separate elements. Initially, if the mobile robot's minimum distance from the obstacle is under $l_a$, it's classified as a collision, resulting in the agent receiving a reward of -500 from the environment. Secondly, the current distance between the mobile robot and the subgoal designed by the high-level policy is less than $l_b$, marked as a successful subgoal reaching, rewarding the agent with 100. Additionally, we define a metric called $d$ that represents the difference in distance to the subgoal between the previous time step and the current time step for the mobile robot. If $d$ is greater than 0, it means that the mobile robot is getting closer to the subgoal, and the environment will return a positive reward of $20d$ to the mobile robot.
Otherwise, if $d$ is less than 0, it indicates that the mobile robot becomes further and further away from the subgoal, and the environment will give a negative reward of -8 to the mobile robot. Up to this, the low-level reward function $R^{l}$ can be expressed as follows:
\begin{align}
R^{l}=\begin{cases} 
  -500& \text{ if collision} \\
  100& \text{ if subgoal reached} \\
  20d& \text{ if $d>0$} \\
  -8& \text{ if $d\leq0$} 
  \label{e:e36}.
\end{cases}
\end{align}

\subsection{Network structure of low-level controller}
\begin{figure}[h]
    \centering
    \includegraphics[width=1\linewidth]{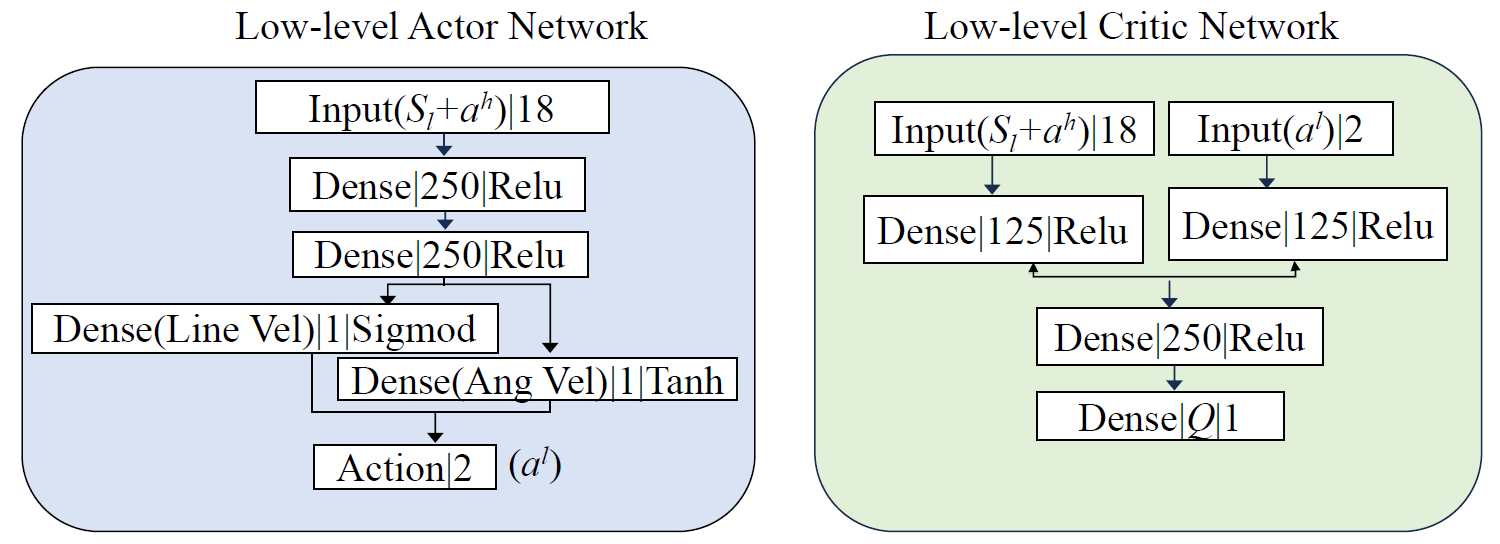}
    \caption{The structure of the actor and critic networks in the low-level controller. }
    \label{fig:network}
\end{figure}
As an actor-critic framework, the DDPG algorithm is designed to integrate both actor and critic networks, each serving a distinct but interconnected purpose in the learning process. The actor network is responsible for determining the optimal policy by mapping states to actions. Regarding to autonomous maze navigation for mobile robots, the low-level actor network inputs the subgoal designed by the high-level policy, along with a state input with 16 dimensions expressed in Eq. \eqref{e:25}. 
The low-level actor network ultimately produces an action output, which includes linear velocity and angular velocity, serving as commands to control the motion of the mobile robot. To ensure the angular velocity remains within the range (-2, 2), the hyperbolic tangent (tanh) function is used as the activation function, effectively constraining the output within the desired limits. Similarly, the sigmoid function is applied as the activation function for linear velocity, restricting its range to between 0 and 1. After applying the activation functions, the linear velocity is scaled by the maximum linear velocity $l_{max}$, and the angular velocity is scaled by the maximum angular velocity $a_{max}$, to compute the final action executed by the mobile robot.

On the other hand, the low-level critic network evaluates the actions chosen by the actor network under a given state. It provides the estimated $Q$ value of an action-state pair, which represents the expected cumulative reward from performing a specific action given a certain state.
The structure of low-level actor and critic networks is illustrated in Fig. \ref{fig:network}. The high-level controller shares the same network architecture as the low-level controller, but the key difference lies in the inputs to its actor and critic networks. The high-level actor network receives a 4-dimensional input, consisting of the $x$, $y$ coordinates of the mobile robot's current position and the 
$x$, $y$ coordinates of the final target destination. In contrast, the high-level critic network takes a 6-dimensional input: the first 4 dimensions correspond to the input information provided to the high-level actor network, while the remaining 2 dimensions represent the high-level action input, which corresponds to the $x$, $y$ coordinates of the subgoal generated by the high-level actor network.

The actor and critic networks update their weights through back-propagation, where gradients of the loss function are propagated backward. Proper weight initialization is crucial for stable training. Large initial weights can cause gradient explosion, leading to unstable or failed training, while excessively small weights can result in vanishing gradients, slowing or halting learning. Both issues hinder the networks' ability to learn meaningful features, resulting in poor performance. Xavier initialization addresses these challenges by setting the initial weights to maintain consistent variance across layers, preventing gradient explosion. For a layer with $n_i$ input nodes and $n_o$ output nodes, the weights are initialized by Xavier initialization from the uniform distribution illustrated as follows:
 \begin{align}
     W \sim \mathcal{U}\left(-\sqrt{\frac{6}{n_{\text {i }}+n_{\text {o }}}}, \sqrt{\frac{6}{n_{\text {i }}+n_{\text {o }}}}\right),
 \end{align}
where $\mathcal{U}$ represents uniform distribution.

\subsection{High-level State Space, Action Space, and Reward Function Design}

\begin{enumerate}
\item State Space: To enhance generality and reduce reliance on highly specific manual design, the state information for the high-level controller is defined as a 4-dimensional vector, expressed as $S$=[$x_m$, $y_m$, $x_f$, $y_f$], where $x_m$ and $y_m$ represent the $x$ and $y$ coordinates of the mobile robot's current position. $x_f$, $y_f$ refer to the $x$ and $y$ coordinates of the final target destination. This comprehensive state representation enables the high-level controller to effectively capture both the mobile robot's current status and its overall navigation goal. By focusing on these critical spatial coordinates, the controller simplifies decision-making, reduces computational complexity, and eliminates the need for overly intricate manual design, thereby improving the algorithm's adaptability. Moreover, this state structure ensures that the generated subgoals are contextually aligned with the mobile robot's position and its ultimate objective, enhancing the efficiency and precision of path planning.
\item Action Space: The actions generated by the high-level policy, which are referred to as subgoals, serve to guide the mobile robot toward the final destination. These subgoals are strategically selected as intermediate targets that are more accessible, simplifying the path to the ultimate goal. To ensure alignment with the final target, the $x$ and $y$ coordinates of the subgoal are directly used to define the high-level actions, maintaining consistency and clarity in representation.
 The high-level action space is expressed as $A^{h}$=[$a^h_x$, $a^h_y$], where $a^h_x$, $a^h_y$ represent the $x$ and $y$ coordinates of the subgoal.

Furthermore, the subgoals generated by the high-level controller should be appropriately constrained in terms of the distance to the current position of the mobile robot to ensure their relevance and feasibility within the mobile robot's operating environment. By limiting the range of potential subgoals, the controller can avoid generating targets that are too far away, unreachable, or outside the robot's working area. This restriction helps maintain the contextual relevance of the subgoals to the mobile robot's current state and its ultimate destination, preventing inefficient or impractical navigation paths. Additionally, position constraints enhance the stability and robustness of the navigation process, as they guide the mobile robot toward intermediate objectives that are both meaningful and achievable, thereby improving overall task efficiency and success rates. Through the experimental trial, we restrict the subgoal to be generated within a 1-meter radius of the mobile robot's current position. 
Additionally, the most fundamental constraint for these subgoals is that they must not go beyond the outer boundary of the maze environment.
\item  Reward Function: The high-level policy prioritizes safe navigation while ensuring path efficiency. Consequently, the high-level reward function must effectively evaluate the quality of the subgoals provided to the low-level controller. Specifically, the high-level controller receives a substantial positive reward of 1000 when the mobile robot successfully reaches the final target destination. Additionally, the reward function assesses whether a subgoal is attainable by the robot. If a collision occurs while attempting to reach a subgoal, the high-level controller incurs a significant penalty of -500. Furthermore, the cumulative rewards obtained by the low-level controller under the guidance of a specific subgoal 
$a^h_t$ are incorporated into the high-level reward function, and a weight adjustment factor $\kappa$ is employed to this sum for stability, set as 0.4 in this work. This integration enables the high-level controller to better perceive environmental dynamics and generate more adaptable subgoals. Accordingly, we design the high-level reward function $R^{h}$ as follows: 
\end{enumerate}
\begin{align}
R^{h}&=r^{h}+{\kappa}\cdot\sum_{i=t}^{i=t+c}R^{l}_i\\
r^{h}&=\begin{cases} 
  -500& \text{ if collision} \\
  1000& \text{ if final goal reached} \\
\end{cases}
\end{align}

\begin{table*}[h]
\begin{center}
\caption{\centering{Experimental hyper-parameters}}
{\fontsize{9}{4}\selectfont
 \begin{tabular} {|c|c|c|}
 \hline
  Parameters& Values &Remark\\
 \hline
 action\_d   & 2    &Dimensions of actions\\
 state\_d &   14  &State space dimension\\
 $\gamma$& 0.99 & Discount factor\\
  M&   200000  & Size of the experience replay memory \\
 BATCH\_SIZE& 256 & Size of sampled batch\\
 
 $\tau$& 0.005 & Update magnitude of target network\\
Max\_Steps\_S1& 300  & Maximum steps for scenario 1 per episode \\
Max\_Steps\_S2& 600  & Maximum steps for scenario 2 per episode \\
Max\_Steps\_S3& 1000  & Maximum steps for scenario 3 per episode \\
  $\eta$& 0.001 & Learning rate \\
 $l_a$& 0.13 m & Threshold of collision \\
 $l_b$& 0.2 m & Threshold of reaching the goal \\
$l_{max}$& 0.22 m/s & The maximum threshold of linear velocity \\
$a_{max}$& 1 rad/s & The maximum threshold of Rotational \\
$\sigma$& 0.2 & Initial standard deviation of the parameter space noise  \\
$\varsigma$& 0.2 & Threshold of the adaptive standard deviation in parameter space noise  \\
$\alpha $& 1.01 & Update frequency of the standard deviation in parameter space noise \\
 \hline
\end{tabular}
}
 \label{t:parameters}
 \end{center}
\end{table*}

\begin{figure*}[htbp]
	\begin{subfigure}{0.3\linewidth}
		\centering
\includegraphics[width=5cm,height=5cm]{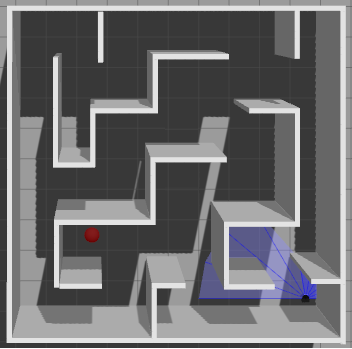}
\captionsetup{justification=centering}
		\caption{Maze scenario 1 with final target of (-2.5, 2.5)}
		\label{fig:stage1}
	\end{subfigure}
    \hfill
 	\begin{subfigure}{0.3\linewidth}
		\centering
		\includegraphics[width=5cm,height=5cm]{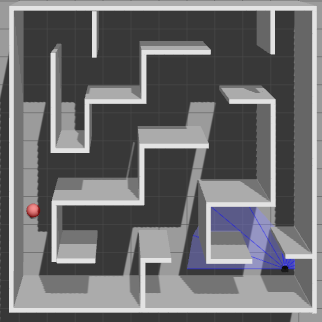}
       \captionsetup{justification=centering}
		\caption{Maze scenario 2 with final target of  (-2.5, 4.5)}
		\label{fig:stage2}
	\end{subfigure}
    \hfill
    \begin{subfigure}{0.3\linewidth}
		\centering
		\includegraphics[width=5cm,height=5cm]{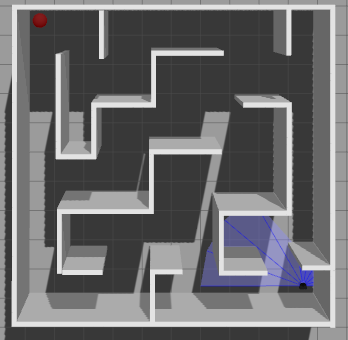}
       \captionsetup{justification=centering}
		\caption{Maze scenario 3 with final target of (4.5, 4.5)}
		\label{fig:stage3}
	\end{subfigure}
        \centering
	\caption{The maze features interconnected rooms and corridors separated by walls, forming a complex 
    layout, where the black point represents the mobile robot, and the red sphere represents the final target.}
	\label{fig:stages}
   \end{figure*}

\section{Numerical Experiments and Results}
\subsection{Experimental Setup}
To evaluate the performance of the proposed algorithm, we conducted extensive experiments simulating autonomous maze navigation tasks for mobile robots, targeting three different final locations. These results were compared with the DDPG algorithm and its variants D$_4$PG. For specific code and videos, you can visit the website\footnote{The code and videos are available at \url{https://github.com/iAerialRobo}}. All experiments were performed on a laptop featuring an NVIDIA GTX4060 GPU (8GB) and 16GB of RAM, employing the open-source Robot Operating System (ROS) and the 3D simulation environment Gazebo. The algorithms were implemented by Python 3.8.10 and PyTorch 1.11, with the experimental specific parameters summarized in Table \ref{t:parameters}.

For a more comprehensive evaluation, we tested the performance and accuracy of three algorithms to autonomously navigate in a maze-like simulation environment, which consists of a grid of 100 cells, arranged in 10 rows and 10 columns, with each cell size of 1 meter. The tested agent is the Waffle model of the TurtleBot3 mobile robot, which navigates to arrive at three distinct final targets illustrated in Fig. \ref{fig:stages}. Scenario 1, depicted in Fig. \ref{fig:stage1}, presents a much easier and nearer final target of (-2.5, 2.5). Scenario 2, as shown in Fig. \ref{fig:stage2} presents a further final target of (-2.5, 4.5). Scenario 3, as shown in Fig. \ref{fig:stage3} presents a much more complicated final target of (4.5, 4.5). Depending on the navigation task, the maze environment, and the size of the mobile robot, setting subgoals generation approximately 1 meter away from the current position of the mobile robot strikes an effective balance. This approach minimizes computational overhead while ensuring efficient and reliable navigation. 

\begin{table*}[]
\caption{\centering{Experiment results trained for 400 episodes in scenario 1 (a maximum of 500 steps per episode)}}
\resizebox{\linewidth}{!}{   

\begin{tabular}{|c|c|c|c|c|c|c|c|c|c|c|c|c|}
\cline{1-13}
Algorithm                  & Indicator & Trail1  & Trail2  & Trail3  & Trail4  & Trail5  & Trail6  & Trail7  & Trail8  & Trail9  & Trail10 & Average \\ \hline
\multirow{2}{*}{DDPG}       & SR  &2.63\% &1.02\% &0.75\% &0.51\% &0.56\% &0.25\% &0.27\% &	0.28\% &0.58\% &0.64\% &0.75\% \\
                           & AS        &-468.60	&-707.65	&-833.34	&-970.41	&-881.05	&-970.24	&-967.74	&-900.76	&-830.64	&-816.28 &-834.67 \\  
\multirow{2}{*}{D$_4$PG}      & SR &32.93\% &33.36\% &32.56\% &27.90\% & 	29.00\% &33.66\% &37.58\% &33.43\% &38.79\% &33.88\% &33.31\%\\	
                           & AS       &294.37	&315.39	&307.84	&264.76	&329.14	&303.34	&360.27	&348.92	&242.84	&278.44&304.53 \\
\multirow{2}{*}{HDDPG} & SR        &88.11\% &90.21\% &89.42\% &90.31\% &91.91\% &89.50\% &91.28\% &88.40\% &90.98\% &88.91\% &89.90\% \\	

                           & AS      	&715.90	&745.69	&866.24	&778.36&791.56	&973.63	&820.49	&840.00	&778.96	&922.76&823.56	 \\ \hline
                         
\end{tabular}
}
\label{t:environment1}
\end{table*}

\begin{table*}[]
\caption{\centering{Experiment results trained for 700 episodes in scenario 2 (a maximum of 600 steps per episode)}}
\resizebox{\linewidth}{!}{   

\begin{tabular}{|c|c|c|c|c|c|c|c|c|c|c|c|c|}
\cline{1-13}
Algorithm                  & Indicator & Trail1  & Trail2  & Trail3  & Trail4  & Trail5  & Trail6  & Trail7  & Trail8  & Trail9  & Trail10 & Average \\ \hline
\multirow{2}{*}{DDPG}       & SR   & 0& 	0.015\% &0.015\% &0& 0& 0& 0& 0&	0&	0.016\% &0.0045\% \\

& AS  &-1431.56	&-1482.67	&-1537.02	&-1599.86	&-1661.73	&-1717.19	&-1663.44	&-1724.40	&-1642.77	&-1329.37	&-1579.00 \\\hline	

\multirow{2}{*}{D$_4$PG}     & SR   	&1.138\% &0.69\% &0.466\% &	0.604\% &0.155\% &0.862\% &5.955\% &0.63\% &0.51\% &0.153\% &1.116\%	\\
 & AS         &-287.53	 &-291.15	 &-294.81	 &-298.54	 &-362.33	 &-366.73	 &-371.16	&-375.69	&-381.674	&-392.27&-342.19	 \\\hline
\multirow{2}{*}{HDDPG} & SR   	&80.29\% &83.94\% &81.36\% &83.37\% &84.68\% &80.18\%  &82.43\% &85.46\% &81.21\% &81.363\% &82.43\%\\
 & AS       	&555.32	&687.54	&308.6	&709.69	&315.54	&1160.76&621.78	&763.65	&737.56	&728.35	&658.879\\ \hline

\end{tabular}
}
\label{t:environment2}
\end{table*}

\begin{table*}[]
\caption{\centering{Experiment results trained for 800 episodes in scenario 3 (a maximum of 1000 steps per episode)}}
\resizebox{\linewidth}{!}{   

\begin{tabular}{|c|c|c|c|c|c|c|c|c|c|c|c|c|}
\cline{1-13}
Algorithm                  & Indicator & Trail1  & Trail2  & Trail3  & Trail4  & Trail5  & Trail6  & Trail7  & Trail8  & Trail9  & Trail10 & Average \\ \hline
\multirow{2}{*}{DDPG}       & SR        & 0\% & 0\% & 0\% & 0\% & 0\% &0\% & 0\% & 0\% & 0\% &0\% &0\% \\
& AS         &-2,798.78	 &2,817.20 &-2,836.53 &	-2,855.74 &-2,875.59	 &2,894.74	 &-2,914.12	 &-2,935.2	 &-2,955.28	 &-2,974.93	  &-2885.79 \\\hline
\multirow{2}{*}{D$_4$PG}       & SR        & 0\% & 0\% & 0\% & 0\% & 0\% &0\% & 0\% & 0\% & 0\% &0\% &0\% \\
 & AS       &-1567.08	 &-1467.6	 &-1470.73	 &-1472.59	 &-1465.92 &-1459.45	&-1475.74	&-1463.9	&-1469.65	&-1463.32	&-1477.60\\\hline

\multirow{2}{*}{HDDPG} & SR &71.42\% &71.58\% &72.92\% &70.70\% &71.36\% &70.50\% &70.433\% &68.24\%&69.81\% &71.24\% &70.82\% \\
& AS    & 639.38& 729.74& 621.18& 432.48&473.03& 498.32& 577.52& 627.28& 534.9	& 547.82& 568.17\\ \hline	
                        
\end{tabular}
}
\label{t:environment3}
\end{table*}

\begin{figure*}[htbp]
    \centering
     \begin{subfigure}[c]{0.48\textwidth}
        \includegraphics[width=1.15\linewidth]{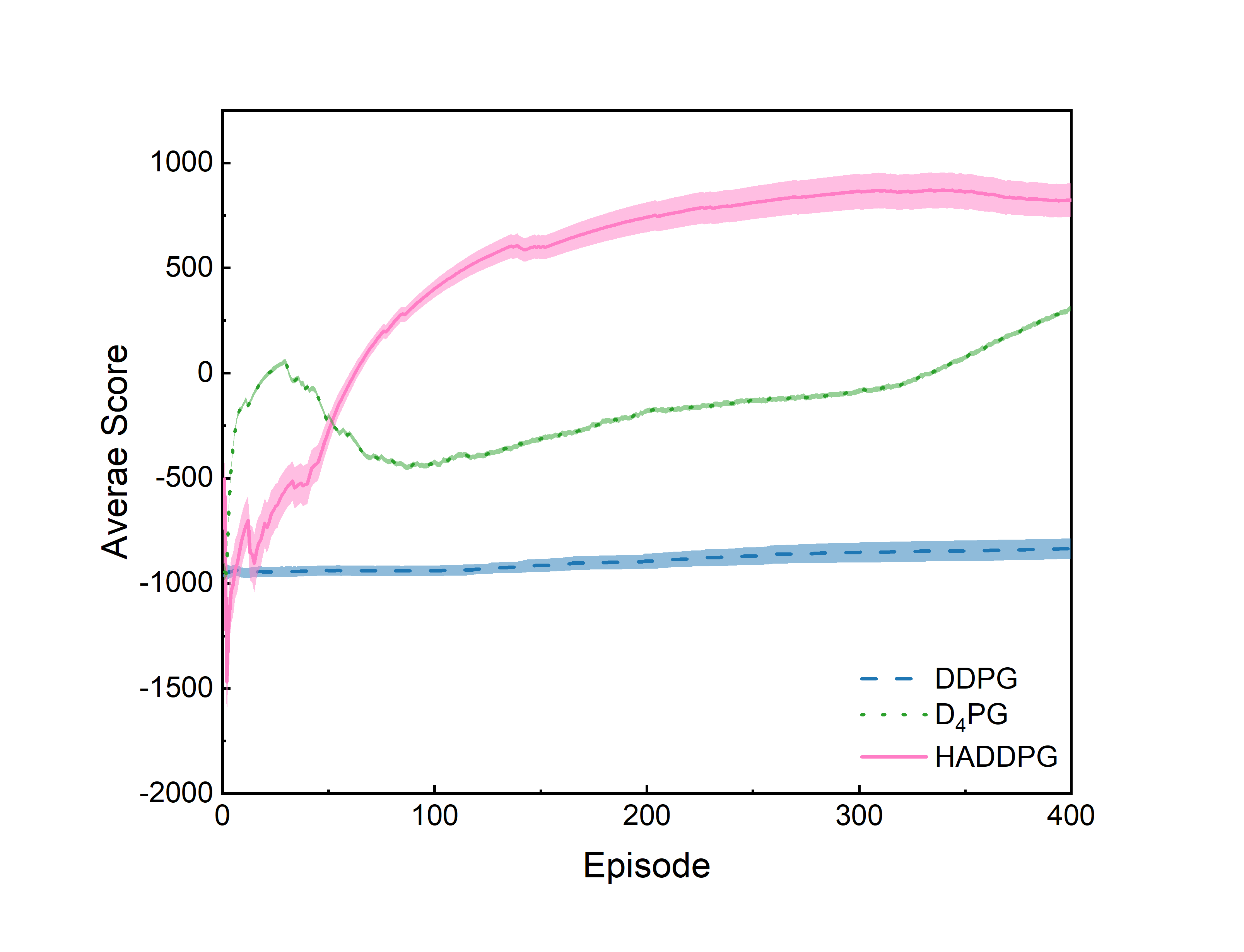}
        \caption{Average score over 10 trials}
        \label{subfig:stage1_as}
    \end{subfigure}
    \begin{subfigure}[c]{0.48\textwidth}
         \includegraphics[width=1.15\linewidth]{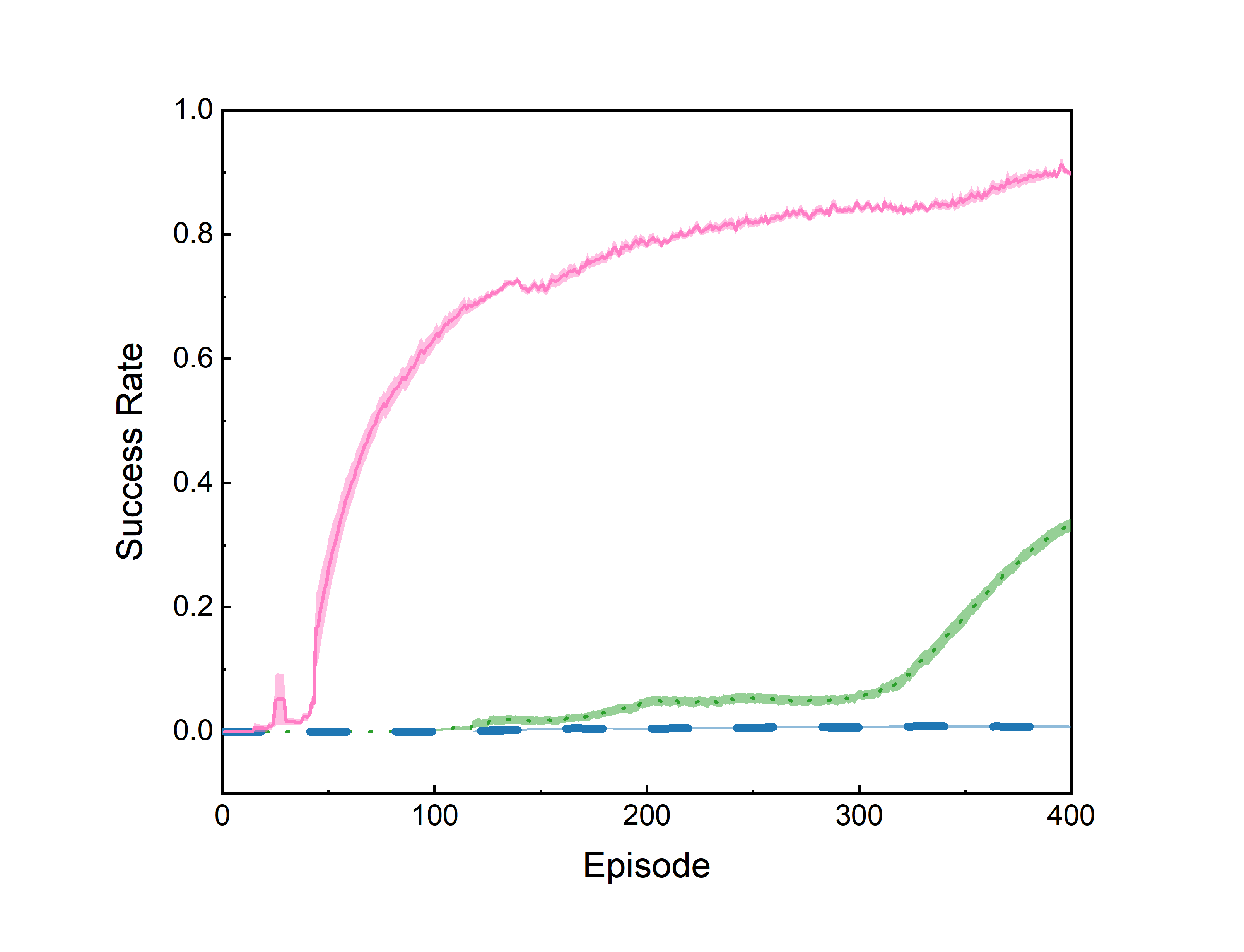}
        \caption{Average success rate over 10 trials}
        \label{subfig:stage1_sr}
    \end{subfigure}
    \caption{Experiment results in scenario 1 trained for 400 episodes (shadowed color: Standard Error)}
    \label{fig:experiment1}
\end{figure*}

\begin{figure*}[htbp]
    \centering
     \begin{subfigure}[c]{0.48\textwidth}
        \includegraphics[width=1.15\linewidth]{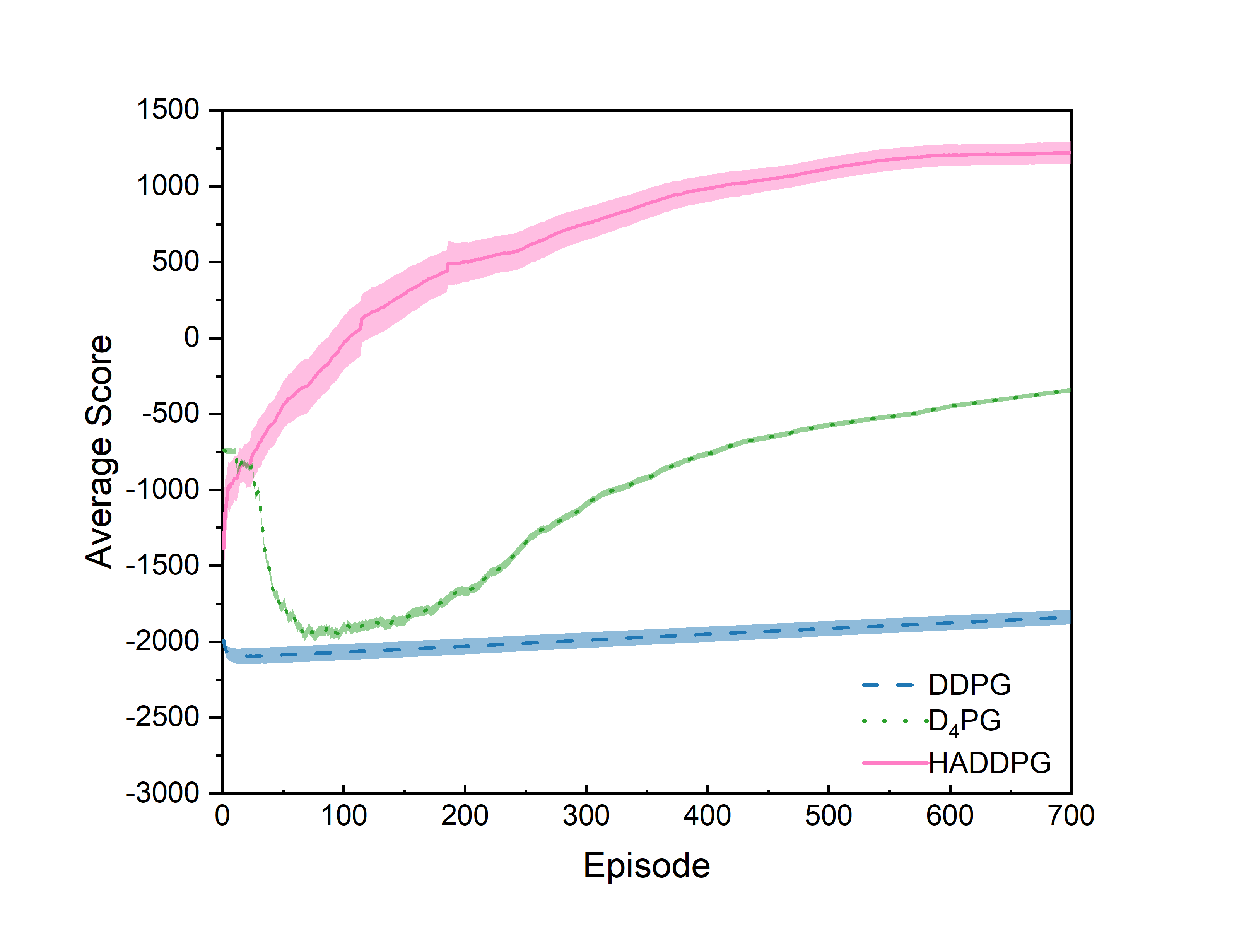}
        \caption{Average score over 10 trials}
        \label{subfig:stage2_as}
    \end{subfigure}
    \begin{subfigure}[c]{0.48\textwidth}
         \includegraphics[width=1.15\linewidth]{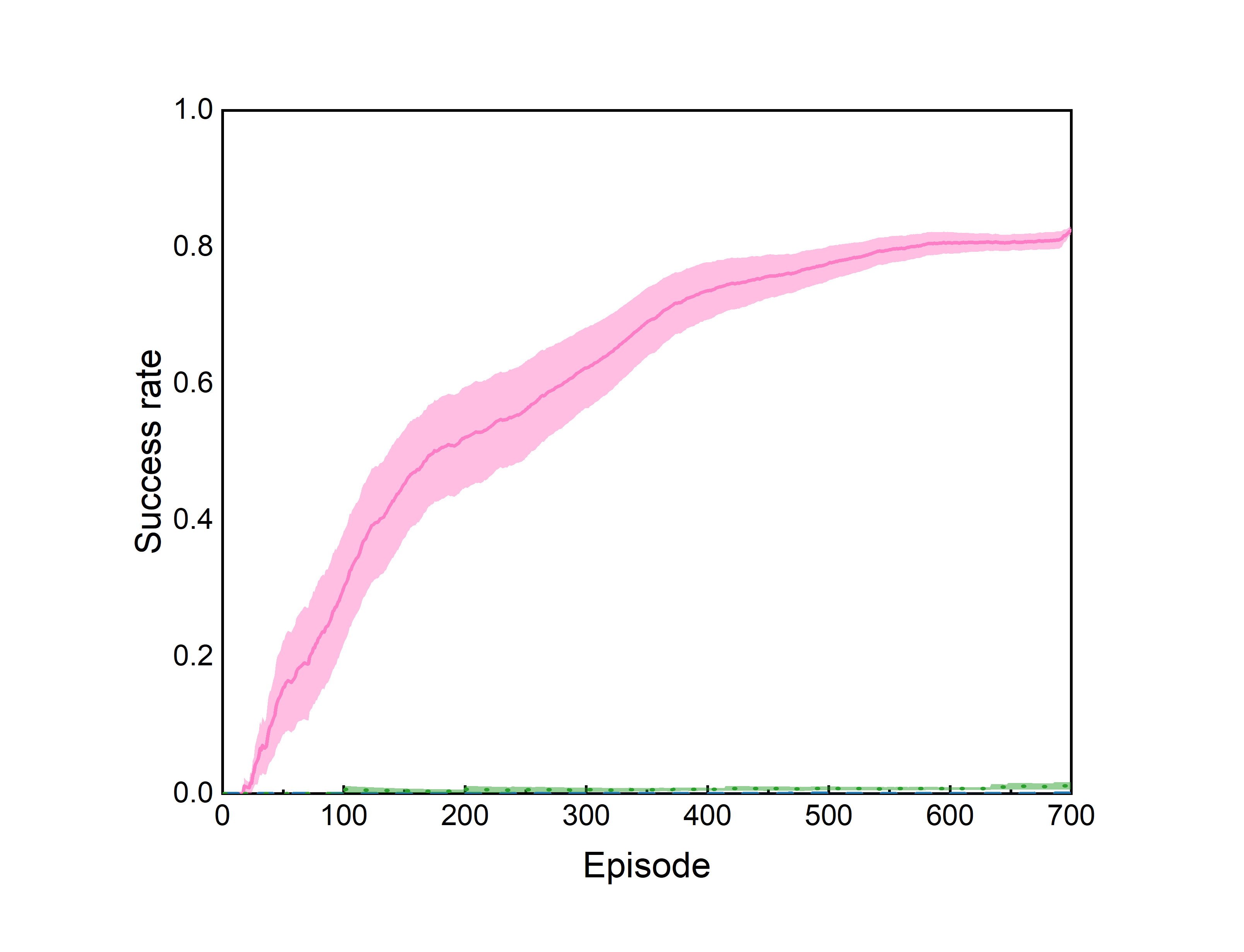}
        \caption{Average success rate over 10 trials}
        \label{subfig:stage2_sr}
    \end{subfigure}
    \caption{Experiment results in scenario 2 trained for 700 episodes  (shadowed color: Standard Error)} 
    \label{fig:experiment2}
\end{figure*}

\begin{figure*}[htbp]
    \centering
     \begin{subfigure}[c]{0.48\textwidth}
        \includegraphics[width=1.15\linewidth]{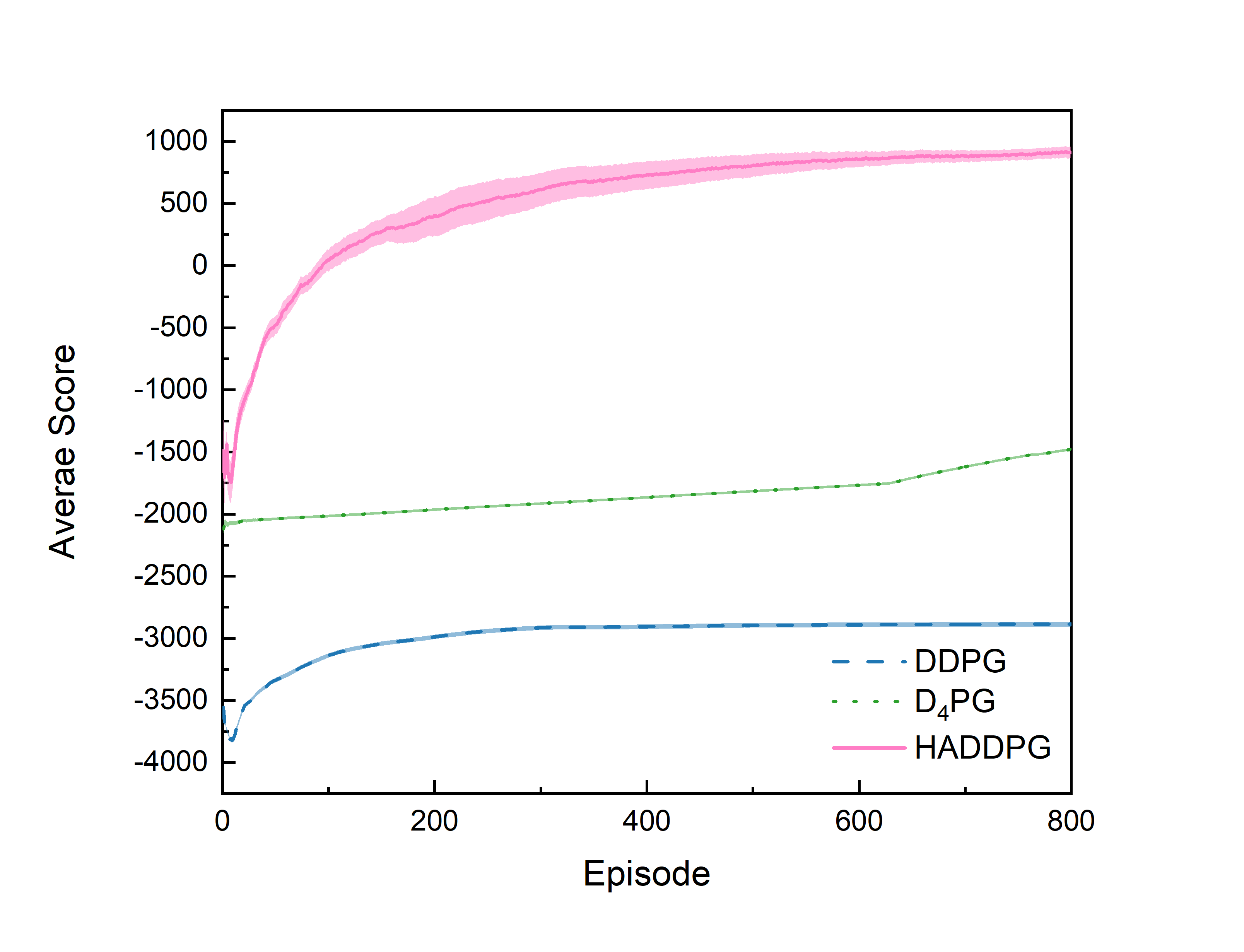}
        \caption{Average score over 10 trials}
        \label{subfig:stage3_as}
    \end{subfigure}
    \begin{subfigure}[c]{0.48\textwidth}
         \includegraphics[width=1.15\linewidth]{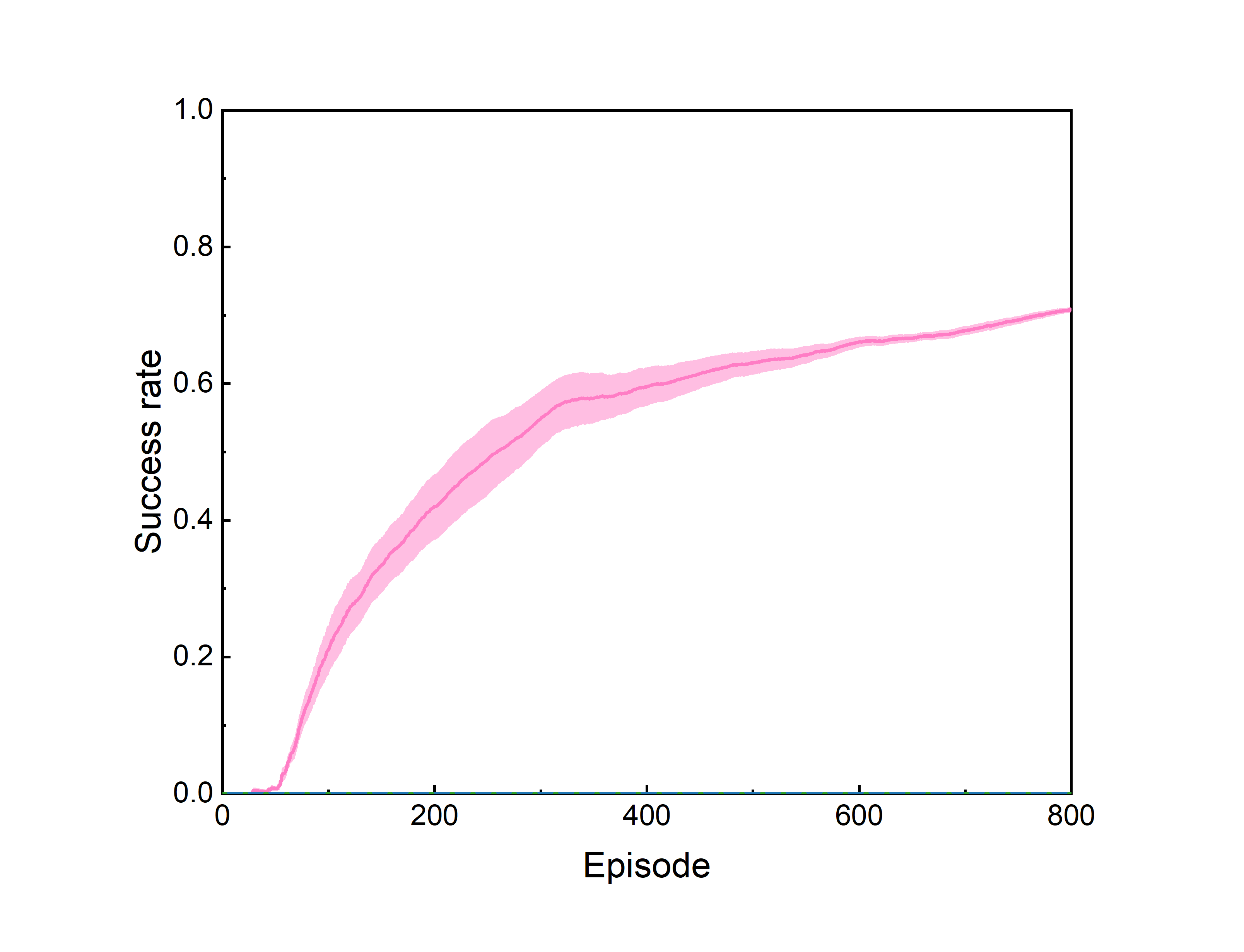}
        \caption{Average success rate over 10 trials}
        \label{subfig:stage3_sr}
    \end{subfigure}
    \caption{Experiment results in scenario 3 trained for 800 episodes  (shadowed color: Standard Error)} 
    \label{fig:experiment3}
\end{figure*}

\begin{figure*}[htbp]
	\begin{subfigure}{0.32\linewidth}
		\centering
\fbox{\includegraphics[width=4cm,height=4cm]{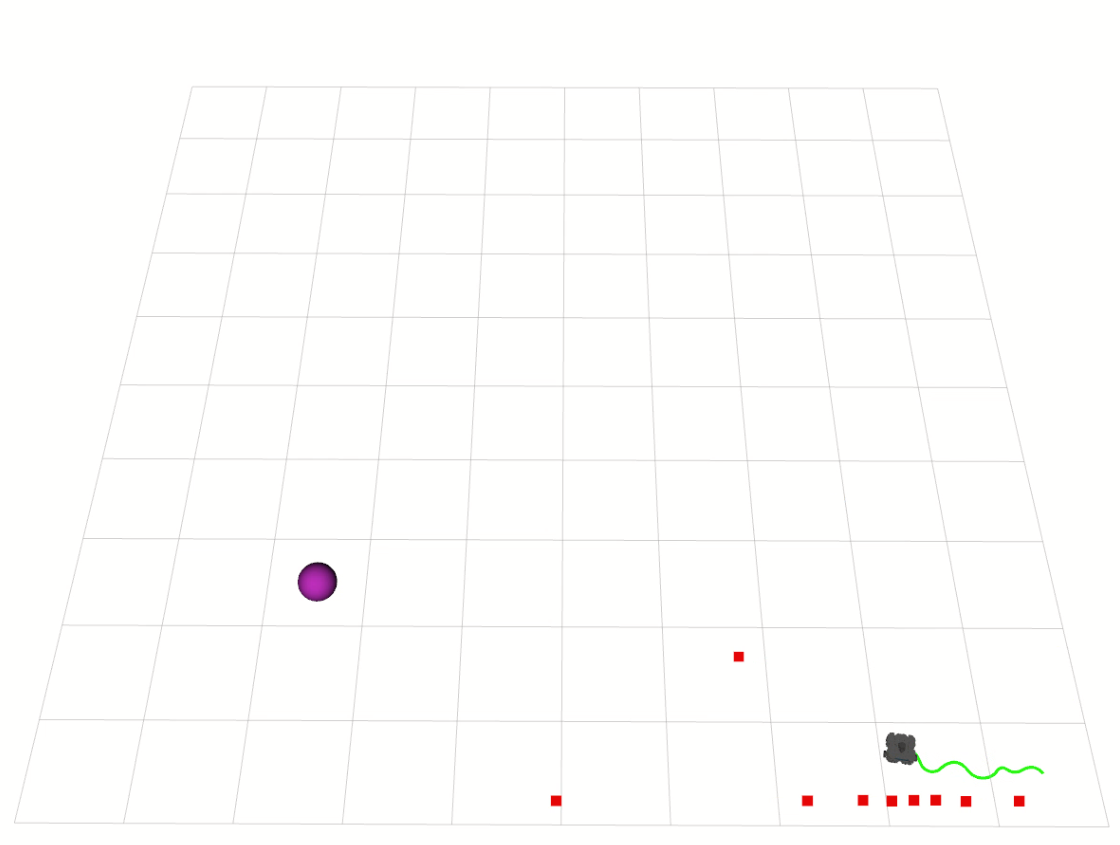}}
\captionsetup{justification=centering}
		\caption{Navigate to the first subgoal}
		\label{fig:stage1_path_1}
	\end{subfigure}
	\centering
	\begin{subfigure}{0.32\linewidth}
		\centering
		\fbox{\includegraphics[width=4cm,height=4cm]{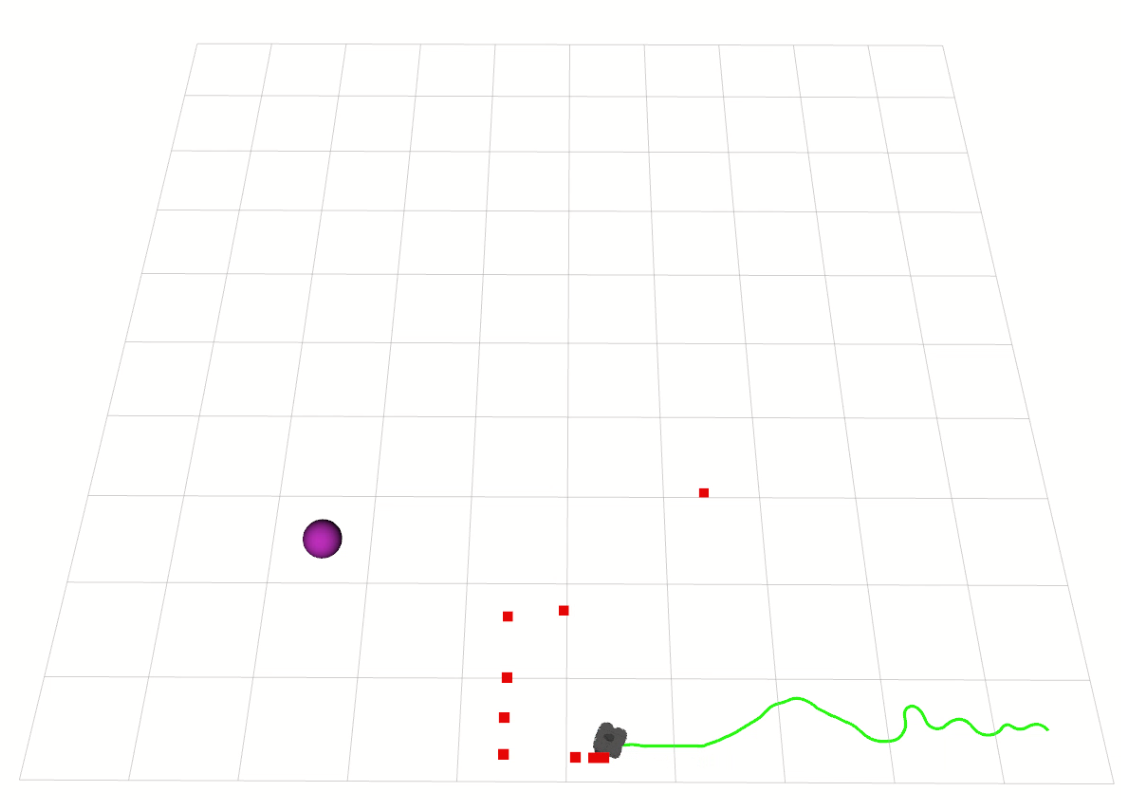}}
       \captionsetup{justification=centering}
		\caption{Navigate to the second subgoal}
		\label{fig:stage1_path_2}
	\end{subfigure}
 	\begin{subfigure}{0.32\linewidth}
		\centering
		\fbox{\includegraphics[width=4cm,height=4cm]{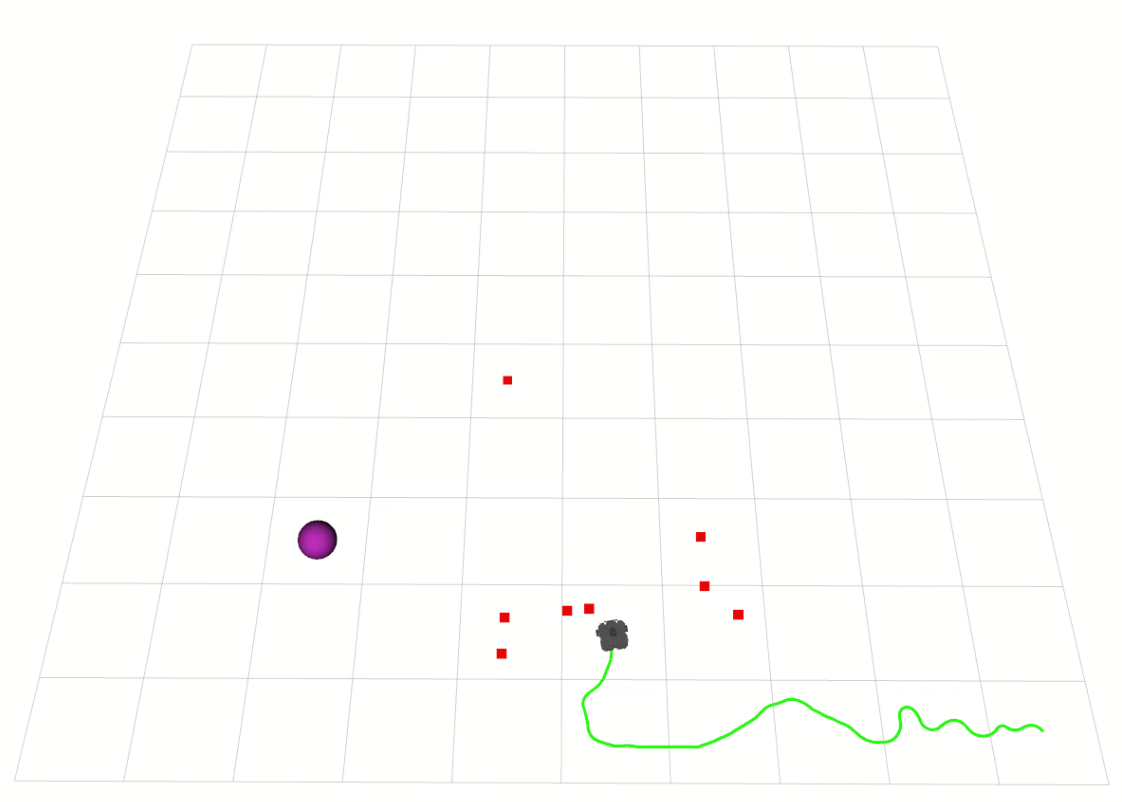}}
       \captionsetup{justification=centering}
		\caption{Navigate to the third subgoal}
        \label{fig:stage1_path_3}
	\end{subfigure}
 \begin{subfigure}{0.33\linewidth}
		\centering
		\fbox{\includegraphics[width=4cm,height=4cm]{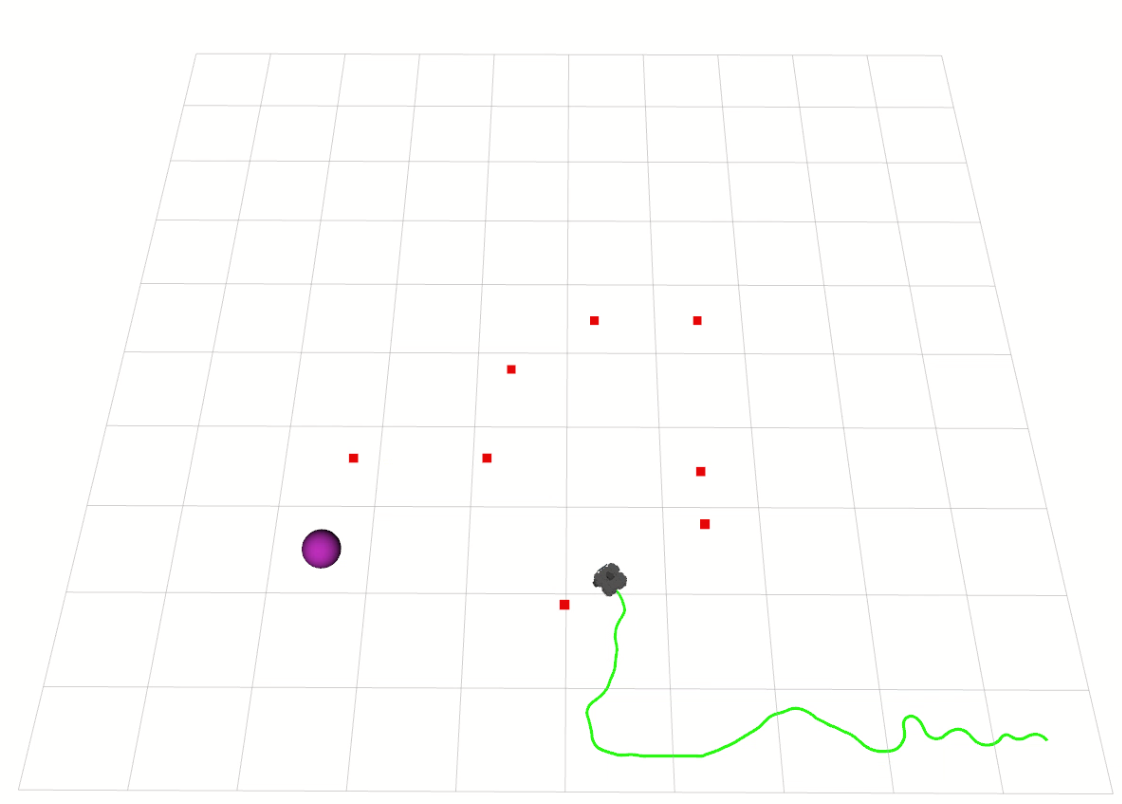}}
       \captionsetup{justification=centering}
		\caption{Navigate to the fourth subgoal}
        \label{fig:stage1_path_4}
	\end{subfigure}
 \begin{subfigure}{0.32\linewidth}
		\centering
		\fbox{\includegraphics[width=4cm,height=4cm]{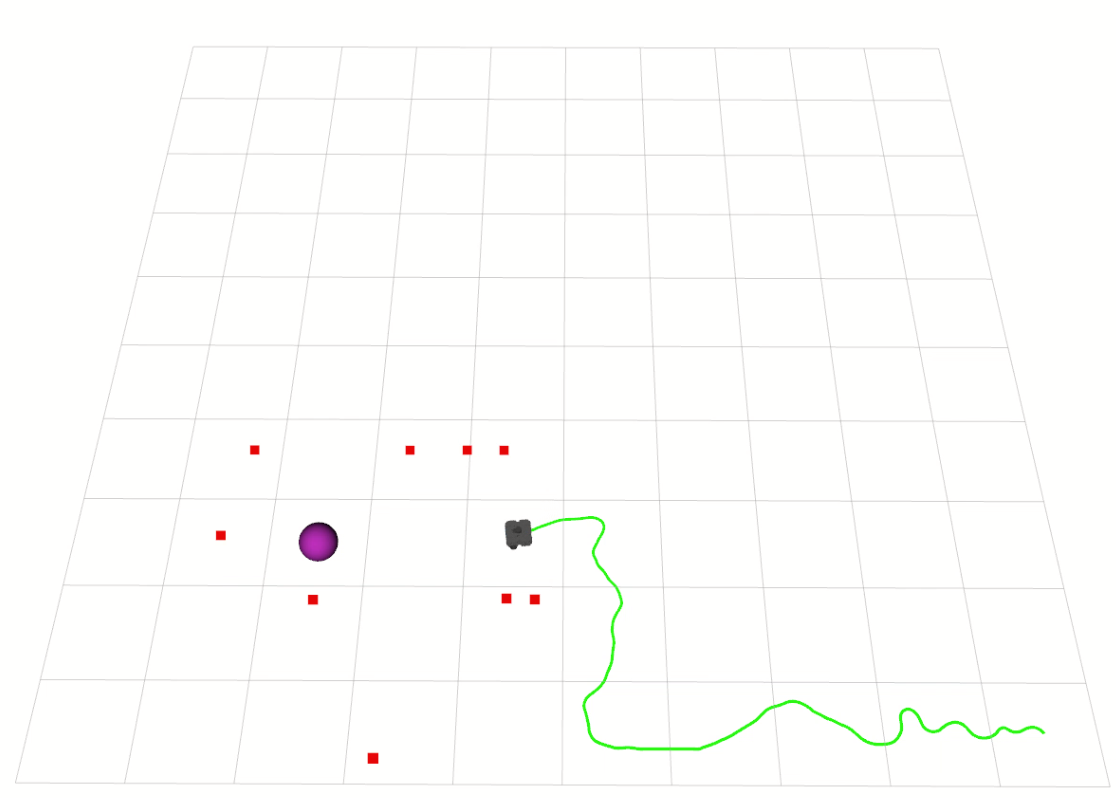}}
       \captionsetup{justification=centering}
		\caption{Navigate to the fifth subgoal}
         \label{fig:stage1_path_}
	\end{subfigure}
 \begin{subfigure}{0.32\linewidth}
		\centering
\fbox{\includegraphics[width=4cm,height=4cm]{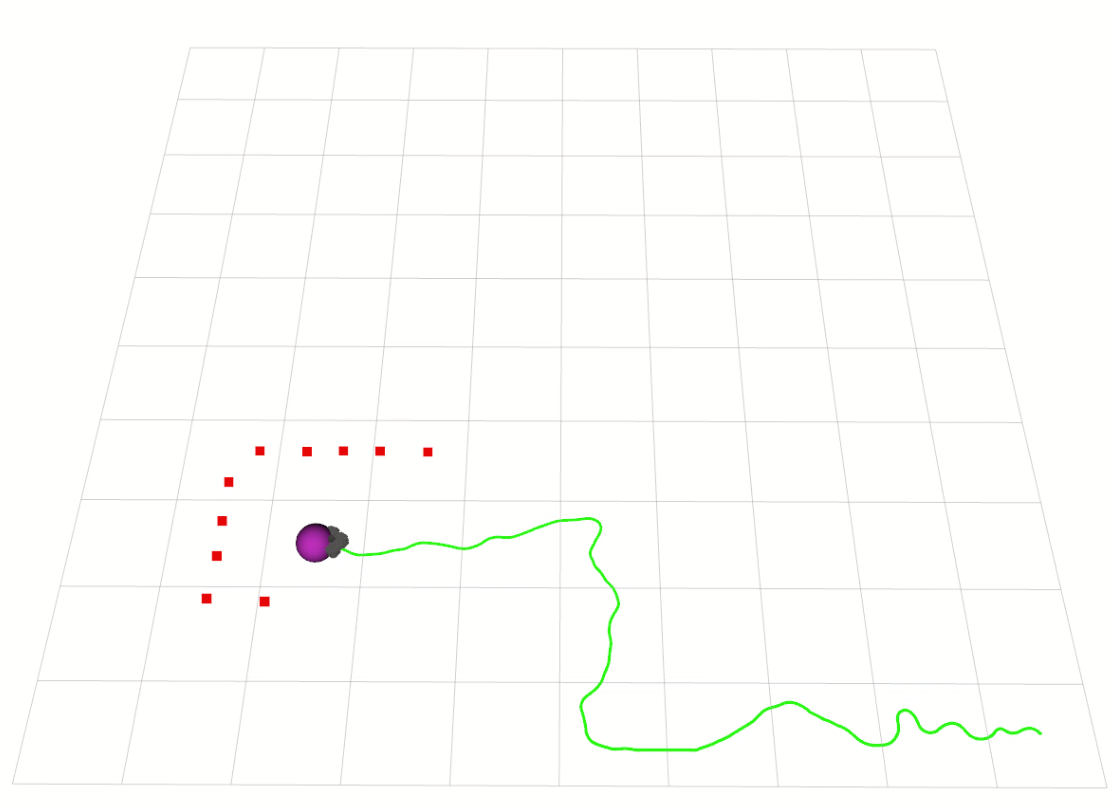}}
       \captionsetup{justification=centering}
		\caption{Navigate to the final target}
        \label{fig:stage1_path_6}
	\end{subfigure}
        \centering
	\caption{The entire path of the robot trained with the HDDPG algorithm in scenario 1 within one episode. The red dots represent the obstacles detected by radar, the purple spheres represent the target points with coordinates (-2.5, 2.5), and the green curves represent the path taken by the robot during the episode. }
   \label{fig:path_stage1}
   \end{figure*}
\begin{figure*}[htbp]
	\begin{subfigure}{0.32\linewidth}
		\centering
\fbox{\includegraphics[width=4cm,height=4cm]{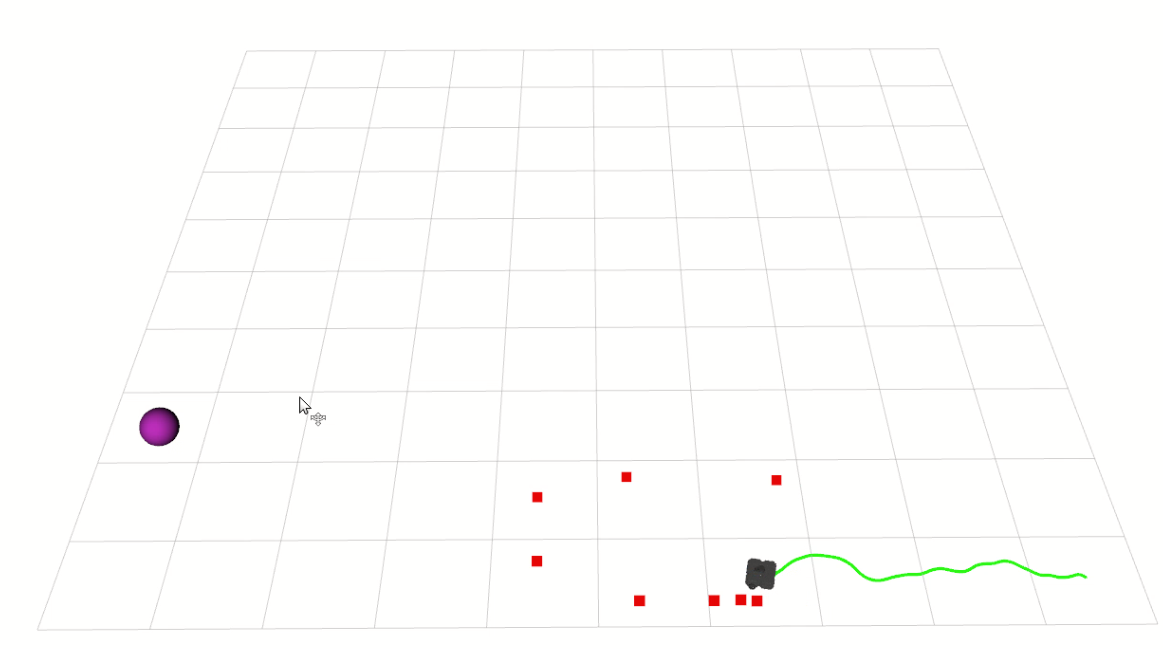}}
\captionsetup{justification=centering}
		\caption{Navigate to the first subgoal}

	\end{subfigure}
     \hfill
	\centering
	\begin{subfigure}{0.32\linewidth}
		\centering
		\fbox{\includegraphics[width=4cm,height=4cm]{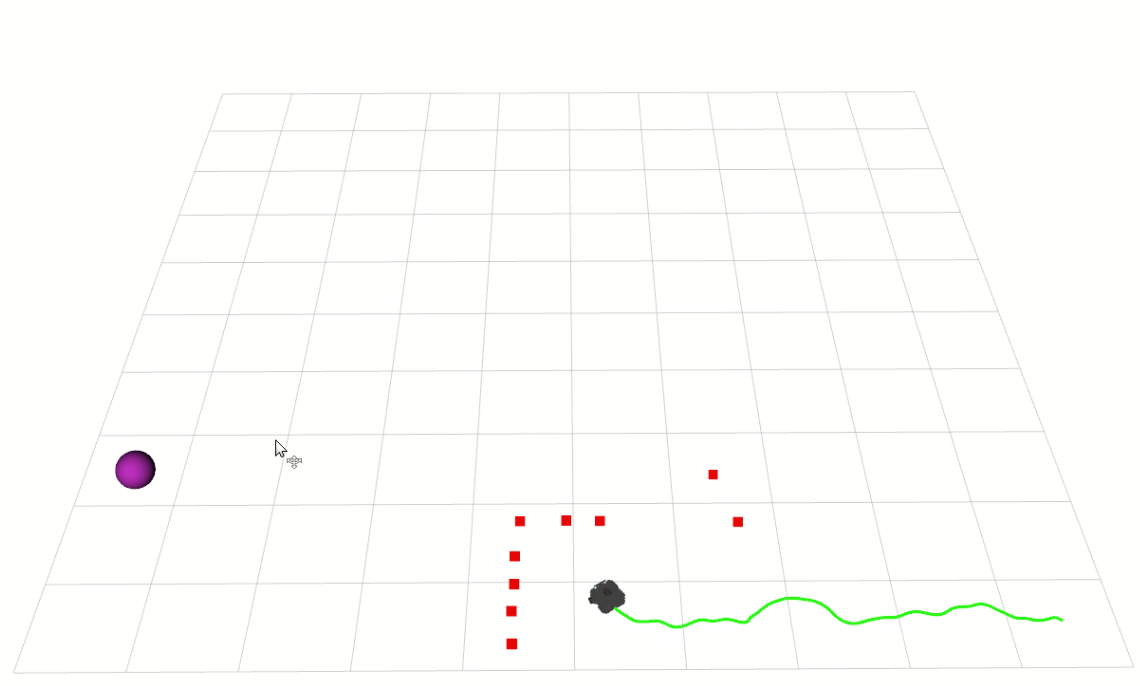}}
       \captionsetup{justification=centering}
		\caption{Navigate to the second subgoal}
	
	\end{subfigure}
     \hfill
 	\begin{subfigure}{0.32\linewidth}
		\centering
		\fbox{\includegraphics[width=4cm,height=4cm]{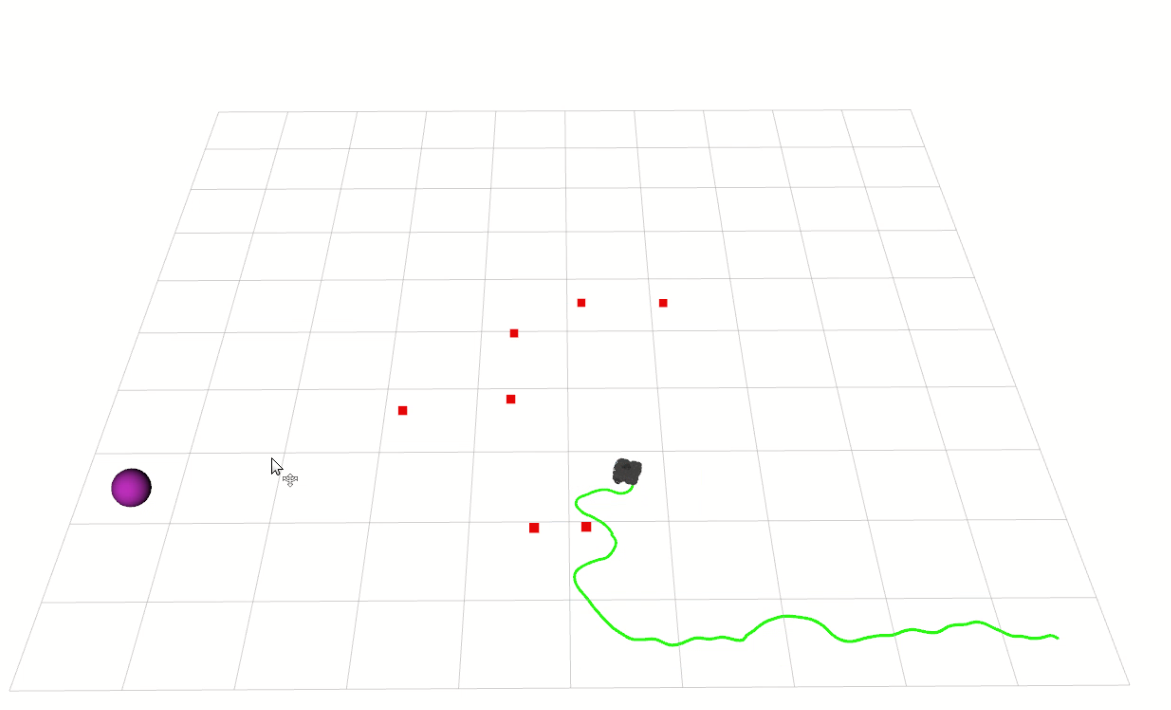}}
       \captionsetup{justification=centering}
		\caption{Navigate to the third subgoal}
	\end{subfigure}
    \vspace{10pt}
 	\begin{subfigure}{0.32\linewidth}
		\centering
		\fbox{\includegraphics[width=4cm,height=4cm]{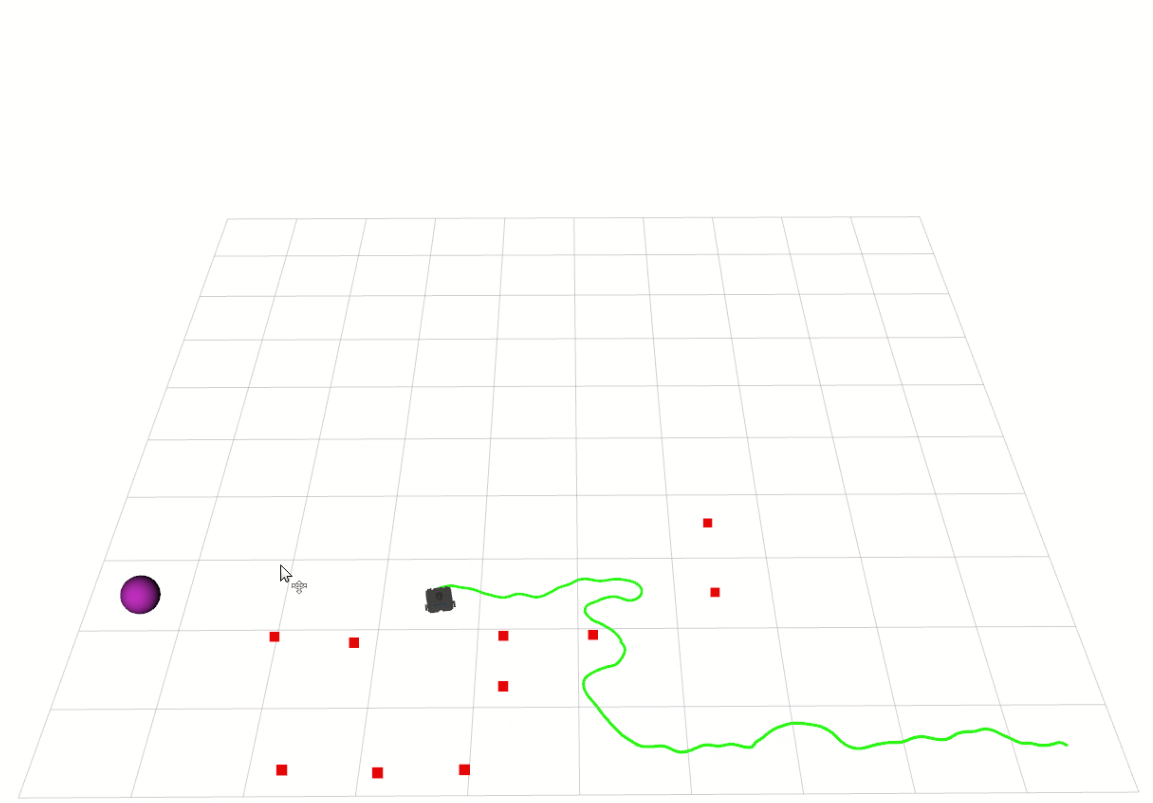}}
       \captionsetup{justification=centering}
		\caption{Navigate to the fourth subgoal}
	
	\end{subfigure}
    \hfill
 	\begin{subfigure}{0.32\linewidth}
		\centering
	\fbox{\includegraphics[width=4cm,height=4cm]{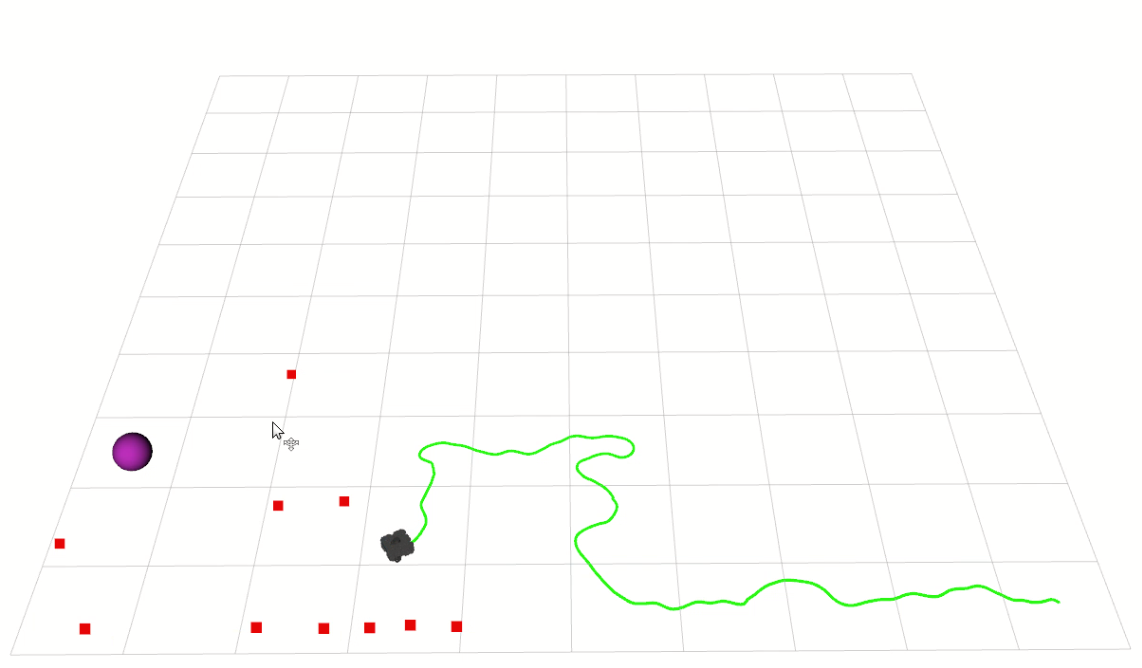}}
       \captionsetup{justification=centering}
		\caption{Navigate to the fifth subgoal}
		
	\end{subfigure}
        \hfill
    \begin{subfigure}{0.32\linewidth}
		\centering
	\fbox{\includegraphics[width=4cm,height=4cm]{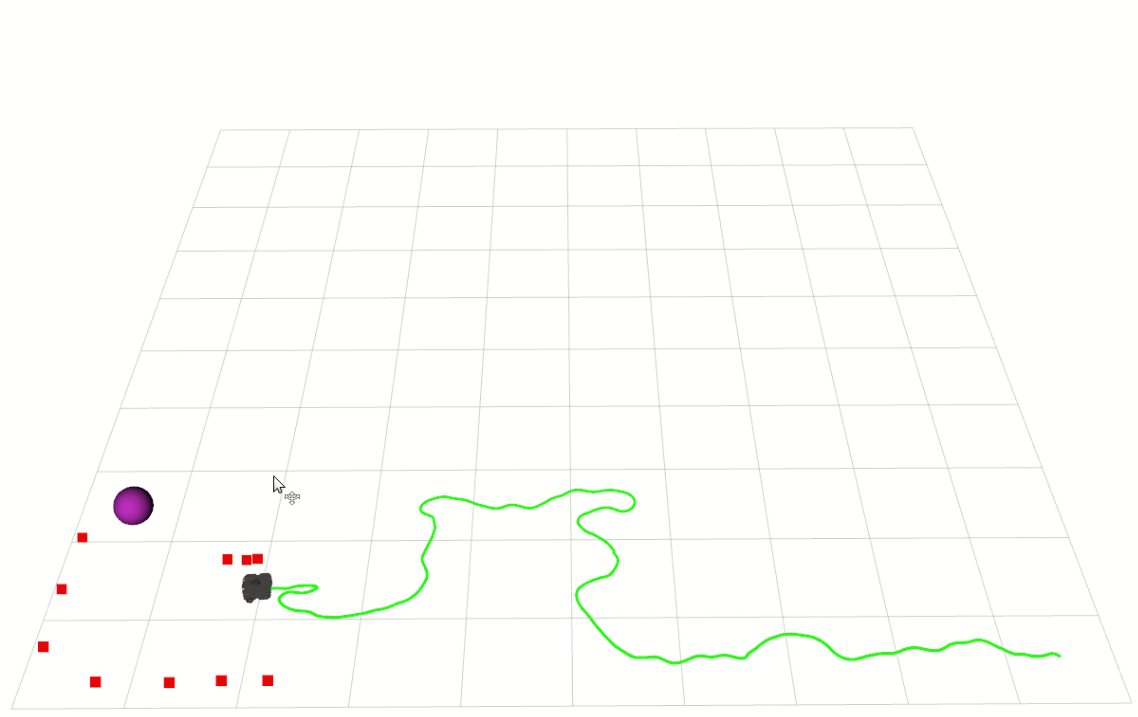}}
       \captionsetup{justification=centering}
		\caption{Navigate to the sixth subgoal}
		
	\end{subfigure}
      \begin{subfigure}{0.32\linewidth}
		\centering
		\fbox{\includegraphics[width=4cm,height=4cm]{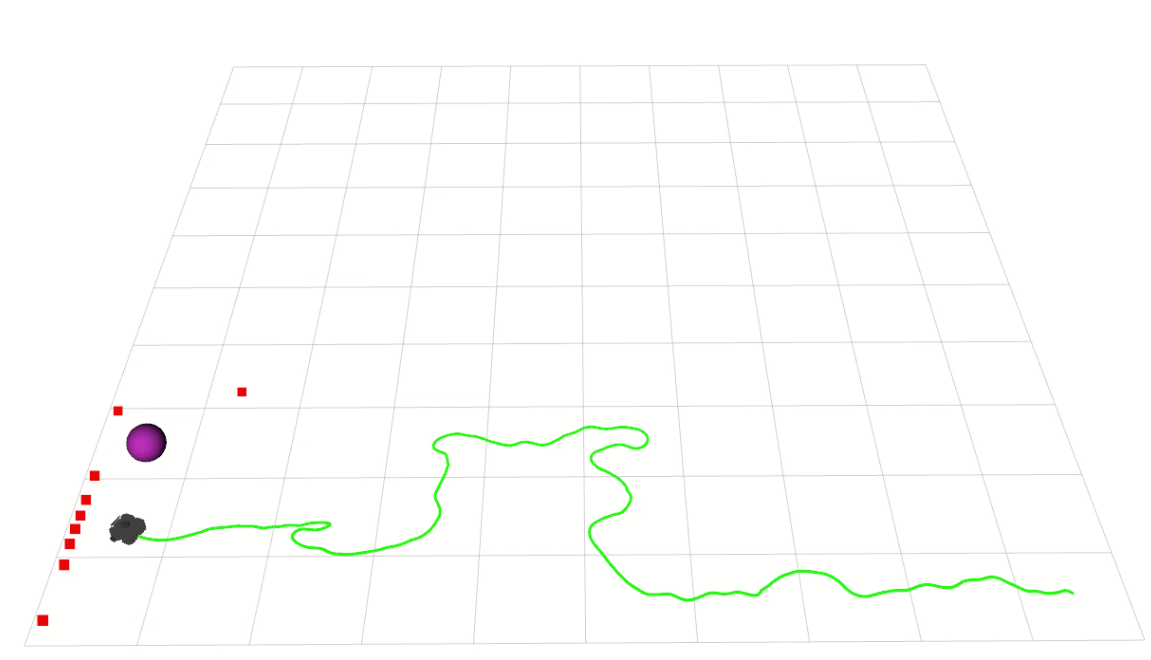}}
       \captionsetup{justification=centering}
		\caption{Navigate to the seventh subgoal}

	\end{subfigure} \hspace{13pt}
    \begin{subfigure}{0.32\linewidth}
		\centering
	\fbox{\includegraphics[width=4cm,height=4cm]{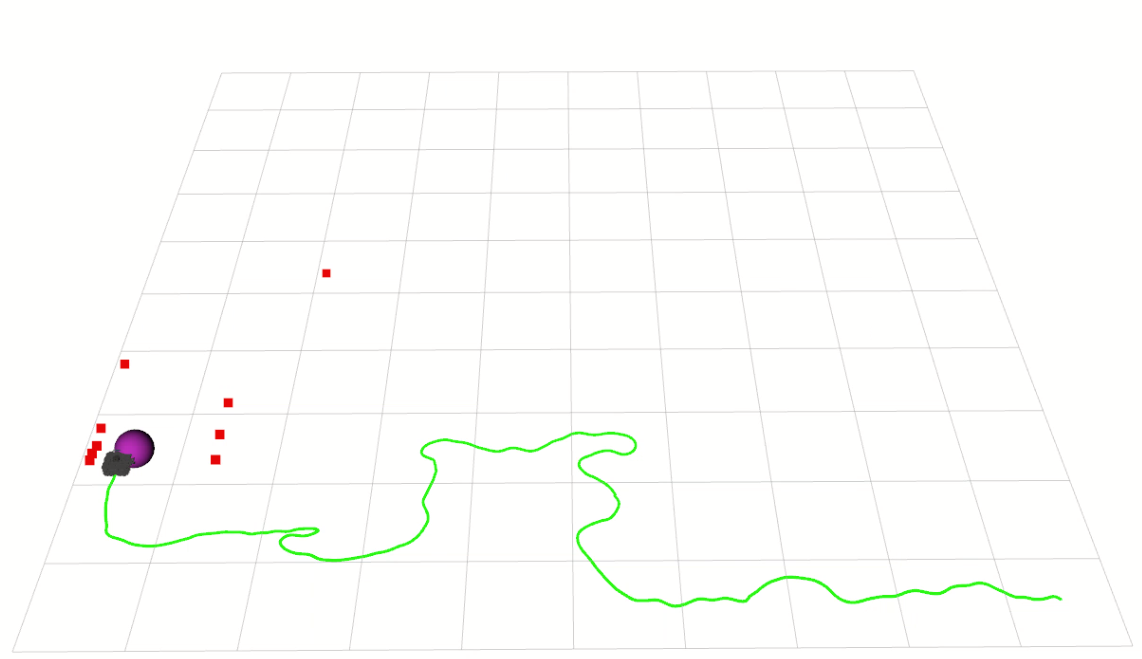}}
       \captionsetup{justification=centering}
		\caption{Navigate to the final target}
	
	\end{subfigure}

        \centering
	\caption{The entire path of the mobile robot trained with the HDDPG algorithm in scenario 2 within one episode. The red squares represent obstacles detected by radar, the purple spheres represent the target point with coordinates (-2.5, 4.5).}
	\label{fig:path_stage2}
   \end{figure*}

\begin{figure*}[htbp]
	\begin{subfigure}{0.32\linewidth}
		\centering
\fbox{\includegraphics[width=4cm,height=4cm]{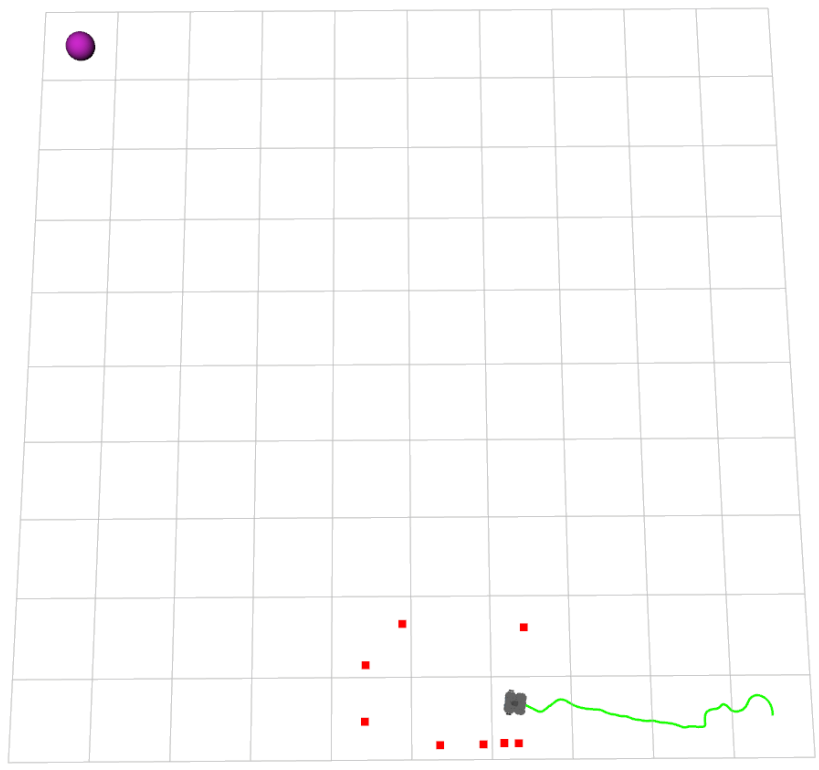}}
\captionsetup{justification=centering}
		\caption{Navigate to the first subgoal}
	
	\end{subfigure}
	\centering
	\begin{subfigure}{0.32\linewidth}
		\centering
		\fbox{\includegraphics[width=4cm,height=4cm]{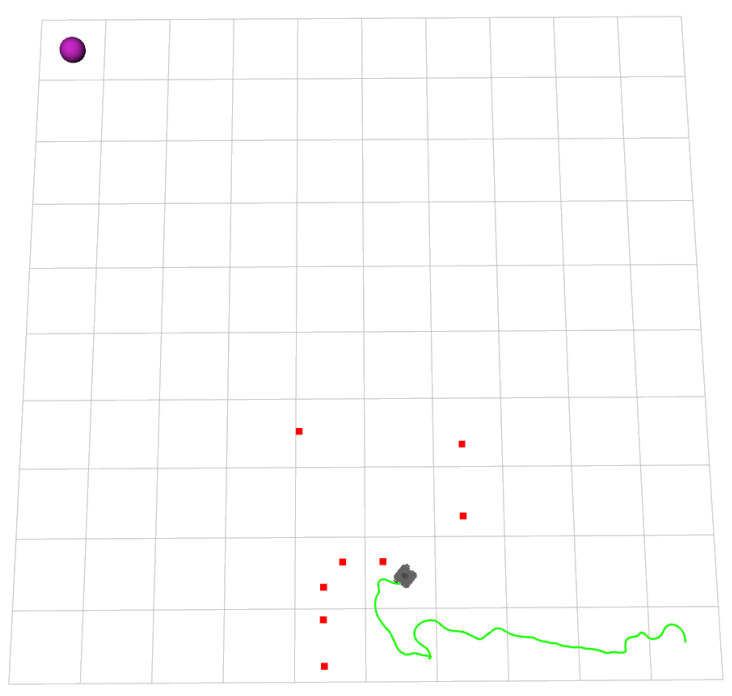}}
       \captionsetup{justification=centering}
		\caption{Navigate to the second subgoal}
	
	\end{subfigure}
 	\begin{subfigure}{0.32\linewidth}
		\centering
	\fbox{\includegraphics[width=4cm,height=4cm]{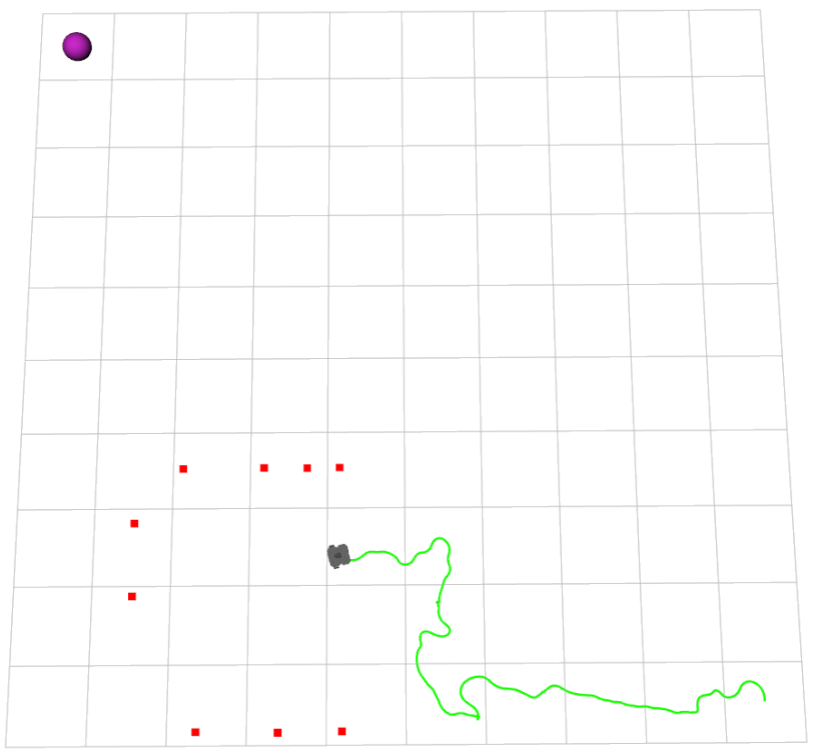}}
       \captionsetup{justification=centering}
		\caption{Navigate to the third subgoal}
	
	\end{subfigure}
 	\begin{subfigure}{0.32\linewidth}
		\centering
		\fbox{\includegraphics[width=4cm,height=4cm]{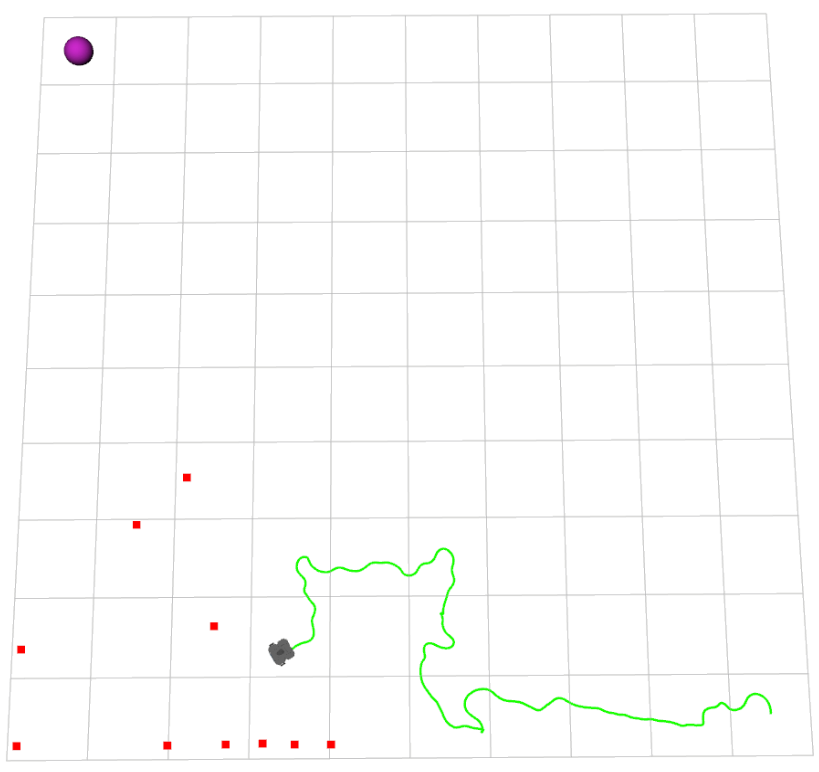}}
       \captionsetup{justification=centering}
		\caption{Navigate to the fourth subgoal}

	\end{subfigure}
 	\begin{subfigure}{0.32\linewidth}
		\centering
		\fbox{\includegraphics[width=4cm,height=4cm]{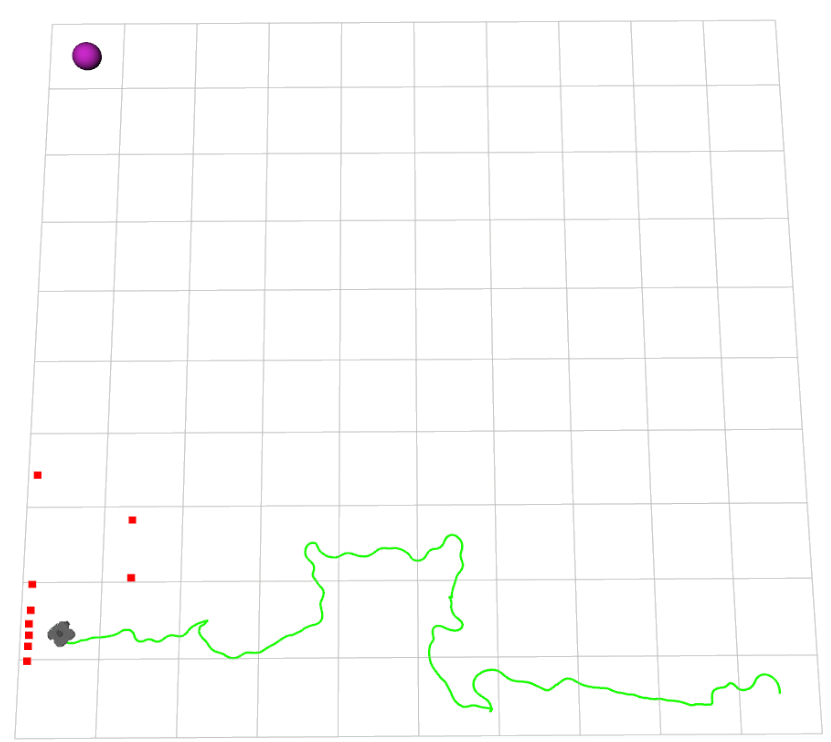}}
       \captionsetup{justification=centering}
		\caption{Navigate to the fifth subgoal}
	
	\end{subfigure}
    \begin{subfigure}{0.32\linewidth}
		\centering
		\fbox{\includegraphics[width=4cm,height=4cm]{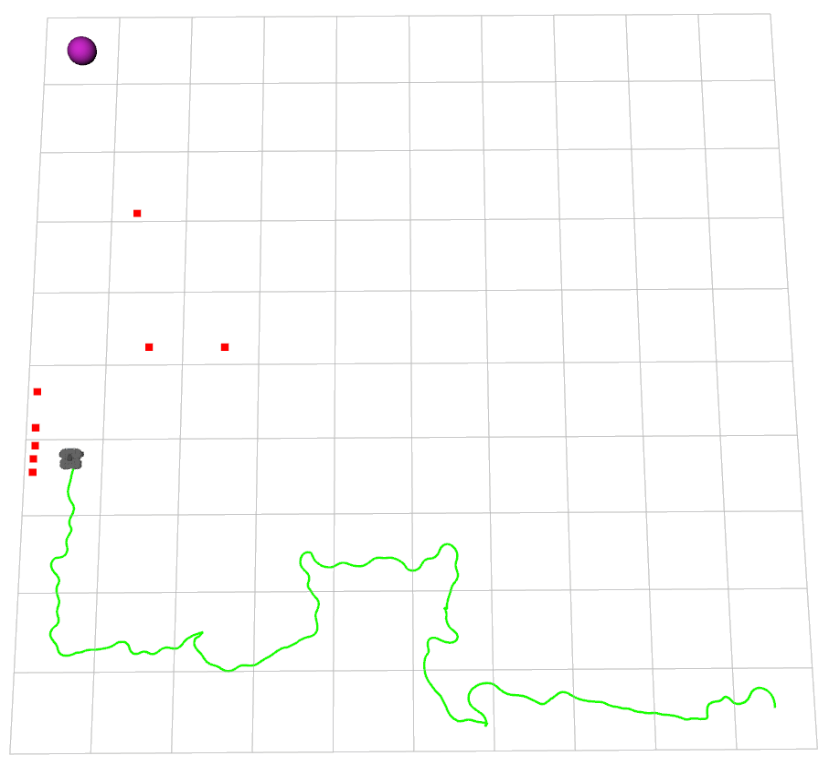}}
       \captionsetup{justification=centering}
		\caption{Navigate to the sixth subgoal}

	\end{subfigure}
    \begin{subfigure}{0.32\linewidth}
		\centering
		\fbox{\includegraphics[width=4cm,height=4cm]{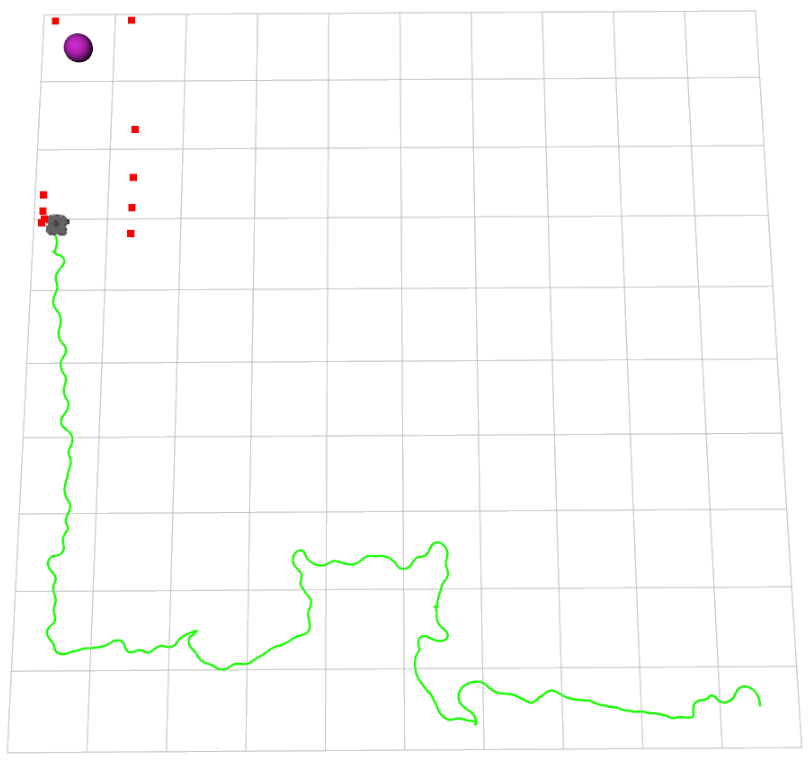}}
       \captionsetup{justification=centering}
		\caption{Navigate to the seventh subgoal }

	\end{subfigure}
    \begin{subfigure}{0.32\linewidth}
		\centering
	\fbox{\includegraphics[width=4cm,height=4cm]{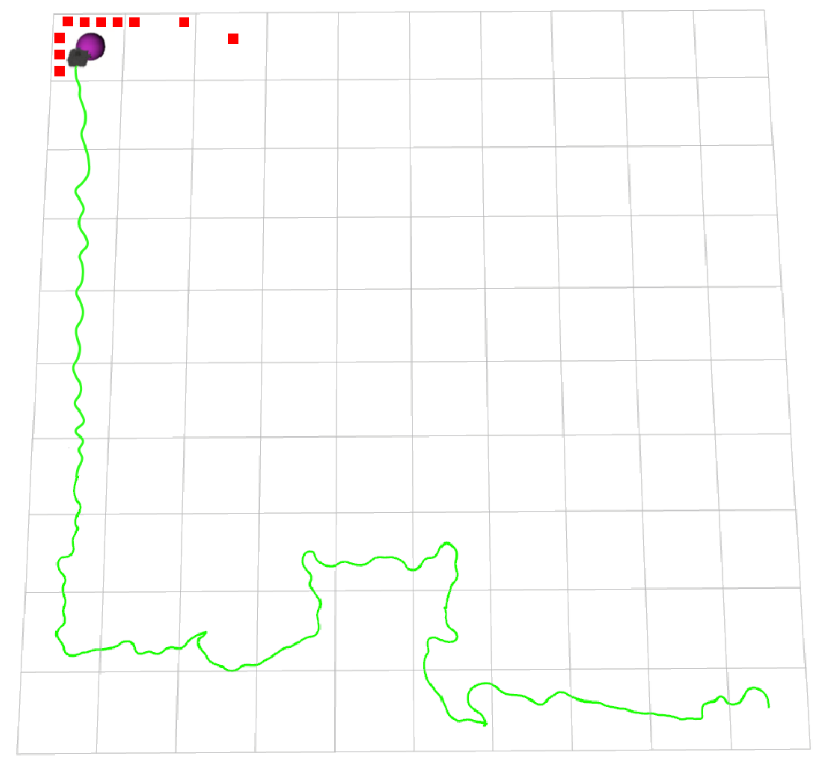}}
       \captionsetup{justification=centering}
		\caption{Navigate to the final target}
	
	\end{subfigure}
        \centering
	\caption{The entire path of the mobile robot trained with the HDDPG algorithm in scenario 3 within one episode. The red squares represent obstacles detected by radar, and the purple spheres represent the target point with coordinates (4.5, 4.5).}
	\label{fig:path_stage3}
   \end{figure*}

\subsection{Results and Discussion}
To evaluate the performance of DDPG, D$_4$PG, and the proposed HDDPG algorithms across three scenarios, this study compares their performance using two metrics: average score (AS) and success rate (SR). 
SR represents success rate. During evaluation we run $N$ independent episodes. An episode is regarded as a “success” if the agent reaches the goal within the maximum allowed number of steps without any collision. SR is calculated as follows:
\begin{align}
SR=\frac{N_{success}}{N},
\end{align}
where $N_{success}$ is the number of episodes in which the agent successfully reaches the goal, and $N$ is the total number of evaluation episodes. AS represents the average score, which is defined as the mean of the flat cumulative rewards obtained by the agent over a fixed number of evaluation episodes. For each episode, the flat cumulative reward is computed by summing the flat rewards received from the initial state until the episode terminates either by reaching the goal or exceeding the maximum number of steps. The AS metric reflects the overall performance of the agent in terms of both task efficiency and flat reward maximization, which is calculated as follows:
\begin{align}
AS &= \sum_{i=1}^{N}R_i, \label{e:e36a} \\
R_i &= \sum_{t=1}^{M}r_t, \label{e:e36b} \\
r_t &= \begin{cases} 
  -500 & \text{if collision} \\
  100 & \text{if subgoal reached} \\
  20d & \text{if $d_g>0$} \\
  -8 & \text{if $d_g\leq0$} 
\end{cases} \label{e:e36c}
\end{align}
where $R_i$ is the cumulative flat rewards for the $i$-th evaluation episode, and $N$ is the total number of evaluation episodes. $r_t$ represents the flat reward at time step $t$. $M$ represents the terminal time step. $d_g$ represents the difference in distance to the final goal between the previous time step and the current time step for the mobile robot.

Fig. \ref{subfig:stage1_as} and \ref{subfig:stage1_sr} depict the performance of three algorithms after 10 training trials in scenario 1 involving an easier and closer target of (-2.5, 2.5), where each trial comprises 400 episodes, and each episode has a maximum of 500 time steps. Table \ref{t:environment1} presents detailed statistics on AS and SR. For SR, DDPG shows poor performance, with an SR of only 0.75\%, indicating its inability to complete tasks effectively. D$_4$PG shows a better improvement, achieving a SR of 33.31\%, which represents a relative improvement of approximately 32. 56\% compared to DDPG. However, D$_4$PG's performance is still unsatisfactory, as it lacks the consistency and reliability needed for robust task completion. On the other hand, HDDPG achieves an outstanding average SR of 89.90\%, a relative of approximately 56.59\% compared to D$_4$PG and 89.15\% compared to DDPG. Its SR ranges consistently high, from 88.11\% to 91.91\%, demonstrating its ability to complete tasks with exceptional success and stability. Regarding AS, DDPG performs the worst, with a negative average score of -834.67, reflecting inefficiency and poor task performance. D$_4$PG shows moderate improvement, with an AS of 304.53, marking a relative improvement of 1139.2 compared to DDPG. However, D$_4$PG’s performance remains suboptimal, as its scores are still far from ideal. HDDPG achieves an exceptional average AS of 823.56, representing a relative improvement of 519.03 compared to D$_4$PG and 1658.23 compared to DDPG. Its AS values range from 715.90 to 973.63, consistently maintaining high scores in all tests. HDDPG demonstrates the best performance both in terms of SR and AS, showcasing its superior consistency and effectiveness in completing maze tasks.

Then, we assessed the performance of the three algorithms in scenario 2, which involves a more distant target of (-2.5, 4.5), and run 10 trials. Each trial consists of 700 episodes, each episode allowing a maximum of 600 time steps. Fig. \ref{subfig:stage2_as} and \ref{subfig:stage2_sr} illustrate the results for AS and SR, respectively, while Table \ref{t:environment2} provides detailed statistical data.
For SR, DDPG still shows exceptionally poor performance, with an SR of just 0.0045\% across all trials, underscoring its inability to navigate in the maze effectively in scenario 2. D$_4$PG shows a slight improvement, achieving an SR of 1.116\%. However, this still reflects a low success rate and insufficient task performance under these conditions. In contrast, HDDPG excels, achieving an impressive SR of 82.43\%, with values ranging from 80.18\% to 85.46\%. This demonstrates HDDPG's ability to consistently complete the task with a high success rate, significantly outperforming both DDPG and D$_4$PG. Referring to AS, DDPG exhibits the poorest performance, with an AS of -1579.00, reflecting severe inefficiency and poor task execution. Although D$_4$PG shows a better improvement of 1236.81, with an AS of -342.19, it still indicates a reduction in negative performance and is far from optimal. HDDPG delivers outstanding results, with an AS of 658.879, representing a dramatic improvement over both DDPG and D$_4$PG. Its scores range from 315.54 to 1160.76, showcasing consistent efficiency and effectiveness across all trials. HDDPG’s performance marks a significant breakthrough, overcoming the substantial limitations of DDPG and D$_4$PG, and demonstrating its capacity to significantly enhance SR and reliability.

Finally, we compared the performance of the three algorithms in scenario 3, which refers to a further target of (4.5, 4.5), executing 10 trials. Each trial consists of 800 episodes, each episode with a maximum of 1000 time steps. The results of AS and average SR are shown in Fig. \ref{subfig:stage3_as} and \ref{subfig:stage3_sr}, respectively, while Table \ref{t:environment3} presents a comprehensive breakdown of the statistical data. For SR, both DDPG and D$_4$PG perform poorly, with an SR of 0\% in all trials, highlighting their entire inability to complete the maze task effectively in scenario 3. In contrast, HDDPG demonstrates remarkable performance, achieving an SR of 70.82\%, with values ranging from 68.24\% to 72.92\%. This highlights HDDPG's ability to consistently complete tasks with a high level of success rate, significantly outperforming both DDPG and D$_4$PG. Among the evaluated methods, DDPG exhibits the poorest performance in AS, just getting an AS of -2885.79. This result underscores its substantial inefficiency and inability to handle the maze navigation task in scenario 3. In comparison, D$_4$PG demonstrates a modest enhancement, achieving an AS of -1477.60. While this signifies a reduction in negative outcomes, the approach remains far from optimal. Conversely, HDDPG achieves outstanding results, with an AS of 568.17, marking an exceptional improvement over both DDPG and D$_4$PG. Its scores range from 432.48 to 729.74, demonstrating consistent efficiency and effectiveness in all trials. Notably, HDDPG represents a breakthrough by overcoming the 0\% success rate limitation of both DDPG and D$_4$PG. This highlights HDDPG's ability to significantly enhance task performance and reliability. 

Across all three scenarios, DDPG and D$_4$PG exhibit limited capability, achieving only moderate success in Setting 1, where the target is relatively close and the environment is less challenging. However, their performance declines sharply in Settings 2 and 3, primarily due to sparse reward feedback, inefficient exploratory behavior, and the absence of temporal abstraction required for effective long-horizon decision-making. These deficiencies result in consistently low success rates and significantly negative average scores, emphasizing the inherent limitations of flat reinforcement learning algorithms when confronted with complex maze navigation tasks featuring narrow passages and delayed rewards. In the overall performance evaluation of the DDPG, D$_4$PG, and HDDPG algorithms across the three scenarios, it is evident that the HDDPG algorithm delivers superior, more stable, and robust performance in autonomous maze navigation for mobile robots. This is reflected in significant improvements in both AS and SR, demonstrating its reliability across varying conditions. Moreover, HDDPG significantly reduces suboptimal decision-making and enhances exploration-exploitation balance, leading to more efficient path planning and faster task completion. These advantages collectively establish HDDPG as a more effective and scalable solution for autonomous maze navigation tasks.

Additionally, path visualization depicted by HDDPG method across three scenarios within an episode is shown in Figs. \ref{fig:path_stage1}, \ref{fig:path_stage2}, and \ref{fig:path_stage3} respectively. In each episode, the mobile robot is allowed to take up to the maximum of time steps. However, if a collision occurs, the current episode ends.

\section{Conclusion}
The proposed approach, HDDPG, incorporates high-level and low-level controllers through an advanced variant of the DDPG method while integrating adaptive parameter space noise and off-policy correction. The high-level policy generates intermediate subgoals from a longer-horizon perspective and on a higher temporal interval, while the low-level policy focuses on creating detailed motions of the agent interacting with the environment. Additionally, adaptive parameter space noise and an off-policy correction mechanism are incorporated to improve exploration efficiency, value estimation, and optimize sub-goal assignment. Moreover, innovative and effective target-driven reshaping of intrinsic and extrinsic reward functions is utilized to refine learning efficiency and improve the success rate. Furthermore, to enhance stability and robustness, valuable enhancements such as the gradient clipping mechanism and Xavier initialization are employed.

To validate the proposed algorithm, we conducted extensive experiments on the ROS and Gazebo platforms, focusing on autonomous maze navigation tasks for a mobile robot with three distinct final target points. The evaluation metrics, including AS and SR, indicate that HDDPG achieves significant improvements and enhanced robustness. Specifically, HDDPG increases the success rate by at least 56.59\% and improves the average reward by no less than 519.03 compared to the benchmark algorithm. This study addresses the challenges in autonomous maze navigation for mobile robots by introducing an optimized hierarchical DDPG algorithm to improve decision-making efficiency. The proposed approach enhances performance in path efficiency, AS, and SR. 

In future work, we aim to enhance the application of HDRL in maze navigation by addressing more complex and dynamic environments. Additionally, we plan to explore multi-agent scenarios with HDRL algorithms, where multiple robots collaboratively navigate in a shared maze environment to significantly enhance the success rate and average score, coupled with reducing exploration time and boosting convergence speed.

\section*{Acknowledgments}
The corresponding author would like to thank the Malaysian Ministry of Higher Education (MOHE) for providing the Fundamental Research Grant Scheme (FRGS) (Grant number: FRGS/1/2024/TK04/USM/02/1). 

\bibliographystyle{elsarticle-num}
\bibliography{main}
\end{document}